\documentclass[review]{elsarticle}
\usepackage[nolist]{acronym}
\usepackage{amsmath,amssymb,amsfonts}
\usepackage[para,online,flushleft]{threeparttablex}
\usepackage{multirow}
\usepackage{graphicx}
\usepackage{textcomp}
\usepackage{xcolor}

\usepackage{tikz}
\usepackage{pgfplots}
\usepackage{pgf-pie}
\usetikzlibrary{plotmarks}
\pgfplotsset{compat=newest}
\usetikzlibrary{patterns}
\usetikzlibrary{shapes,arrows}

\usepackage{epsfig}
\usepackage{epstopdf}
\usepackage{graphicx}
\usepackage{amssymb}
\usepackage{array}
\usepackage{arydshln}
\usepackage[utf8]{inputenc}
\usepackage{balance}
\usepackage{url}

\usetikzlibrary{external}
\usepackage{pgf-pie}
\usepackage{etoolbox}
\newtoggle{showpct}
\makeatletter
\patchcmd{\pgfpie@slice}%
{\scalefont{#3}\beforenumber#3\afternumber}%
{\iftoggle{showpct}{\scalefont{#3}\beforenumber#3\afternumber}{}}%
{}{}
\makeatother

\usepackage{subfig}

\newlength\figureheight
\newlength\figurewidth

\makeatletter
\def\endthebibliography{%
	\def\@noitemerr{\@latex@warning{Empty `thebibliography' environment}}%
	\endlist
}
\makeatother

\newcolumntype{L}[1]{>{\raggedright\let\newline\\\arraybackslash}m{#1}}
\newcolumntype{C}[1]{>{\centering\let\newline\\\arraybackslash}m{#1}}
\newcolumntype{R}[1]{>{\raggedleft\let\newline\\\arraybackslash}m{#1}}

\newcommand{\totalReviewedPapers}{66}
\newcommand{\pklotOverallUsagePct}{88\%}
\newcommand{\cnrparkOverallUsagePct}{28\%}
\newcommand{\pldsOverallUsagePct}{1\%}
\newcommand{\overallParkingClassificationOnly}{73\%}
\newcommand{\globalAverageAccSinglePK}{96.1\%}
\newcommand{\globalAverageAccCamChange}{92.7\%}
\newcommand{\globalAverageAccPkChange}{91.8\%}
\newcommand{\globalParkingSpotDetectionWorks}{13\%}
\newcommand{\numWorksIndividualClassConsiderChange}{35\%}
\newcommand{\globalCarDetecCountWorks}{14\%}

\usepackage{todonotes}

\begin{document}

\begin{acronym}[TDMA]
    \acro{AMF}{Adaptive Median Filter}
    \acro{AP}{Average Precision}
    \acro{AR}{Average Recall}
    \acro{AUC-ROC}{Area Under the ROC Curve}
    \acro{AUC-PR}{Area Under the Precision-Recall Curve}
    \acro{CNN}{Convolutional Neural Network}
    \acro{CSV}{Comma-separated Values}
    \acro{DL}{Deep Learning}
    \acro{DWT}{Discrete Wavelet Transform}
    \acro{FN}{False Negative}
    \acro{FNR}{False Negative Rate}
    \acro{FP}{False Positive}
    \acro{FPR}{False Positive Rate}
    \acro{GAN}{Generative Adversarial Network}
    \acro{GPS}{Global Positioning Systems}
    \acro{GLCM}{Gray-Level Co-Occurrence Matrix}
    \acro{HOG}{Histogram of Oriented Gradients}
    \acro{HSV}{Hue, Saturation and Value}
    \acro{IoU}{Intersection over Union}
    \acro{k-NN}{k-nearest neighbors}
    \acro{LBP}{Local Binary Patterns}
    \acro{LPQ}{Local Phase Quantization}
    \acro{LR}{Logistic Regression}
    \acro{MAD}{Mean Absolute Deviation}
    \acro{MAE}{Mean Absolute Error}
    \acro{mAP}{Mean Average Precision}
    \acro{MLP}{Multilayer Perceptron}
    \acro{MSD}{Mean Signed Deviation}
    \acro{MSE}{Mean Squared Error}
    \acro{NMS}{Non-Maximum Suppression}
    \acro{PLds}{Parking Lot dataset}
    \acro{QLRBP}{Quaternionic Local Ranking Binary Pattern}
    \acro{RFID}{Radio-Frequency Identification}
    \acro{ROC}{Receiver Operating Characteristic}
    \acro{ROI}{Region of Interest}
    \acro{TN}{True Negative}
    \acro{TP}{True Positive}
    \acro{ULBP}{Uniform Local Binary Pattern}
    \acro{XML}{Extensible Markup Language}
    \acro{RMSE}{Root Mean Squared Error}
    \acro{SURF}{Speeded Up Robust Features}
    \acro{SVM}{Support Vector Machine}
    \acro{WoS}{Web of Science}
\end{acronym}

\journal{Expert Systems with Applications}

\bibliographystyle{model5-names}\biboptions{authoryear}

\begin{frontmatter}

\title{A Systematic Review on Computer Vision-Based Parking Lot Management Applied on Public Datasets}

\author[1]{Paulo~Ricardo~Lisboa~de~Almeida}\corref{correspondingauthor}
\cortext[correspondingauthor]{Corresponding author}
\ead{paulo@inf.ufpr.br}
\author[1]{Jeovane~Honório~Alves}
\ead{jhalves@inf.ufpr.br}
\author[2]{Rafael~Stubs~Parpinelli}
\ead{rafael.parpinelli@udesc.br}
\author[3]{Jean Paul Barddal}
\ead{jean.barddal@ppgia.pucpr.br}

\address[1]{Department of Informatics, 
Federal University of Paran\'{a} (UFPR), Curitiba (PR), Brazil}

\address[2]{Graduate Program in Applied Computing, Santa Catarina State University (UDESC), Joinville (SC), Brazil}

\address[3]{Graduate Program in Informatics (PPGIa), Pontif\'{i}cia Universidade Cat\'{o}lica do Paran\'{a} (PUCPR), Curitiba, Brazil}

\begin{abstract}
Computer vision-based parking lot management methods have been extensively researched upon owing to their flexibility and cost-effectiveness. To evaluate such methods authors often employ publicly available parking lot image datasets. In this study, we surveyed and compared robust publicly available image datasets specifically crafted to test computer vision-based methods for parking lot management approaches and consequently present a systematic and comprehensive review of existing works that employ such datasets. The literature review identified relevant gaps that require further research, such as the requirement of dataset-independent approaches and methods suitable for autonomous detection of position of parking spaces. In addition, we have noticed that several important factors such as the presence of the same cars across consecutive images, have been neglected in most studies, thereby rendering unrealistic assessment protocols. Furthermore, the analysis of the datasets also revealed that certain features that should be present when developing new benchmarks, such as the availability of video sequences and images taken in more diverse conditions, including nighttime and snow, have not been incorporated.
\end{abstract}

\begin{keyword}
Parking lot \sep dataset \sep benchmark \sep machine learning \sep image processing
\end{keyword}

\end{frontmatter}

\section{Introduction}\label{sec:introduction}

Over the last few years, many authors proposed computer vision-based approaches to address problems related to parking lot management.
These problems focus on processing images from parking lots. They include different goals, such as (i) the automatic detection of parking spaces positions, e.g., defining for each parking space the bounding box that delimits the object, (ii) the individual parking space classification, determining whether a specific parking spot is occupied by a vehicle or not, and (iii) detecting and counting vehicles in images.
The tasks mentioned above are often the core components of Smart Parking solutions. 
They aim at providing, among others, automated parking lot management, e.g., dynamic pricing according to the number of cars in the parking lot; and parking guidance for drivers, e.g., a route to the nearest parking space available.
Smart Parking solutions are essential as a system that accurately guides the driver to the nearest parking spot available can save both time and fuel \citep{polycarpouEtAl2013,paidiEtAl2018}.

In this study, publicly available parking lot image datasets were surveyed. In addition, the computer vision-based works that have employed such datasets to address parking lot management problems were evaluated. The scope of this study was limited to computer vision-based approaches as they are advantageous over individual sensors. For example, in contrast to magnetometers and ultrasonic sensors, a single camera can monitor a wide parking area eliminating the need of requiring a sensor per parking spot. Furthermore, cameras reduce installation and maintenance costs and can aid in additional tasks, such as abnormal behavior and theft detection \citep{paidiEtAl2018,liEtAl2019,vargheseSreelekha2019}.

To the best of our knowledge, this work is the first dataset-centered review of computer vision-based approaches to address problems related to parking lot management.
It is relevant to mention that sensors other than cameras may be used for automatic parking lot management. 
Such sensors are beyond the scope of this work. Recent comprehensive reviews of different sensors and approaches for managing parking lots can be found in \citep{polycarpouEtAl2013,fraiferFernstrom2016,paidiEtAl2018,barrigaEtAl2019}.

Reproducibility is among the most important guidelines to be followed in any research. Thus, this review focused on works that utilize at least one robust and publicly available parking lot image dataset, thereby increasing the reproducibility of experiments performed. Moreover, we also propose a criterion that the parking lot image datasets must encompass real-world challenges, avoiding trivial and unrealistic problems. Owing to this filtering process, the selected datasets that fulfilled the this criterion were further described and compared. We expect this review to aid researchers in (i) the development of new computer vision-based parking lot management methods, and (ii) with the proposal of novel robust datasets that can be employed to validate such methods.
Furthermore, we reviewed the existing vision-based approaches, which use at least one of the surveyed datasets, addressing the following problems:

\begin{itemize}
	\item Classification of individual parking spaces;
	\item Automatic detection of parking spaces positions;
	\item Vehicles counting or detection;
\end{itemize}

As the reviewed approaches use publicly available datasets, these are easier to compare and verify. Moreover, the surveyed approaches were also categorized and compared in this work. Consequently, the results obtained were used to identify well-studied solutions for problems such as the individual parking slots classification. In addition, research gaps that researchers could further investigate, for example, the automatic detection of parking spaces and problems generated by camera angle changes, were also identified. The complete contributions of this work are as follows:

\begin{itemize}
    \item We propose certain criteria to define a parking lot image dataset as robust;
    \item We bring forward a review of existing robust parking lot image datasets;
    \item We review, categorize and compare the results of state-of-the-art works that use the surveyed datasets;
    \item After analyzing the state-of-the-art methods and results, we identify the research gaps that researchers should address in the future.
\end{itemize}

This paper is divided as follows.
Section \ref{sec:motivation} brings forward a discussion on existing surveys and their main shortcomings, which are used as guidance for determining our research method.
The research procedure used in this work, including the criteria established to select the public parking lot images studies and datasets, is brought forward in Section \ref{sec:researchMethod}.
Details about the selected datasets (PKLot, CNRPark-EXT, and \acs{PLds}) are given in Section \ref{sec:datasets}. 
By following the research method presented in Section \ref{sec:researchMethod}, we found \totalReviewedPapers\ works that use the datasets mentioned above. 
These works are presented in Section \ref{sec:stateoftheartreview}, where the approaches are categorized according to the following tasks: individual parking spot classification, automatic parking space detection, and car detection and counting.
In Section \ref{sec:discussion}, we summarize and discuss the datasets and the reviewed works. 
Furthermore, we also present our findings, such as that most authors focus on the classification of individual parking spaces.
Finally, the conclusions and envisioned future works are presented in Section \ref{sec:conclusion}.

\section{Motivation} \label{sec:motivation}

In this section, we outline the research method used.
To justify this research method and the scope of our work, we first highlight and discuss nine surveys and reviews that cite the datasets analyzed in this work \citep{meduriEstebanez2018,mahmudEtAL2020,paidiEtAl2018,enriquezEtAl2017,kawade2020survey,barrigaEtAl2019,chenEtAl2019,zantalis2019review,diaz2020survey}.
In \citet{meduriEstebanez2018}, a brief review of \ac{DL} based approaches for parking lots is given. Nevertheless, the authors considered only six works. 
They stated how questionable the generalization of these works is since the samples' conditions, e.g., weather, lightning, occlusions, and car size, are reasonably similar.
The authors also point out that no solution detected the parking spaces automatically.
\citet{mahmudEtAL2020} surveyed some works related to the individual parking spaces classification problem and, similarly to \citet{meduriEstebanez2018}, concluded that generalization issues and automatic parking spot detection are open problems.

Most reviews focus on discussing the different sensors available for individual parking space monitoring and user software, e.g., smartphone applications, developed to manage the parking lots or guide drivers to the parking spaces \citep{paidiEtAl2018,enriquezEtAl2017,kawade2020survey,barrigaEtAl2019}.
The sensors often surveyed in these works include ultrasonic, \ac{RFID}, magnetometers, microwave radars, and camera sensors.

The authors in \citet{paidiEtAl2018} claim that individual parking spaces monitoring in open areas is still an open problem in the literature. In \citet{enriquezEtAl2017} is presented a survey about infrastructure-based, vision-based, and crowd-sensing-based solutions for on- and off-street solutions. 
They argued that vision-based systems have problems with illumination, occlusions, and generalization capabilities. 
On the other hand, infrastructure-based solutions have higher costs. In contrast, crow-sensing solutions require many contributors, which are not always available. 

In \citet{kawade2020survey}, a survey is presented, including hardware, i.e., ultrasonic and infrared sensors, and computer vision-based works. However, only a limited number of papers were analyzed. 
They stated that installation and maintenance were problems found in sensor-based works, as occlusion was in computer vision-based ones. 
\ac{DL}-based approaches for smart cities' problems were surveyed in \citet{chenEtAl2019}. These problems include human mobility, traffic flow, traffic surveillance, and parking lot management.
The authors conclude that, besides its importance, \ac{DL}-based solutions suffer from problems such as high computational cost and the lack of training datasets.
Likewise, \citet{zantalis2019review} surveyed different techniques for smart city problems, including Smart Parking.
In \citet{diaz2020survey}, the authors present a systematic review of different types of Smart Parking systems, such as guidance, reservation, and crowdsourcing.

The works of \citet{enriquezEtAl2017,paidiEtAl2018, barrigaEtAl2019,chenEtAl2019,zantalis2019review, diaz2020survey,kawade2020survey} offer a broader vision of the parking lot management systems, including different sensors and user-level software. 
However, these works do not present an in-depth discussion about vision-based parking lot management approaches. In addition, they did not discuss publicly available datasets developed to verify these approaches.
The only surveys that have focused on vision-based parking lot management problems are \citet{meduriEstebanez2018,mahmudEtAL2020}. 
Nevertheless, these works are not systematic reviews and considered only a few works in the analysis.
Therefore, the systematic mapping of vision-based solutions that can be reproduced using public datasets proposed here has its relevance justified.

\section{Research Procedure}\label{sec:researchMethod}

In this section, we bring forward the research procedure adopted to conduct a systematic literature review on computer vision-based parking lot management systems and public datasets.
First, we have defined the following research questions (RQs) to guide the identification and assessment of relevant works:

\begin{itemize}  
    \item RQ{$_1$}: Which are the primary parking lot management problems dealt with by computer vision-based state-of-the-art solutions?
    \item RQ{$_2$}: What are the primary computer vision-based techniques employed in the state-of-the-art to solve parking lot management problems? 
    \item RQ{$_3$}: What are the open problems in computer vision-based techniques not covered by the state-of-the-art or that still require more research?
\end{itemize}

With the scope defined, we discuss the planning and conduction of the review. We first find the publicly available image parking lot datasets and then collect related works. 
We applied three keywords in the well-known Scopus\footnote{Scopus Website: www.scopus.com (Accessed on Aug 23, 2021).} and \ac{WoS}\footnote{\ac{WoS} Website: www.webofknowledge.com (Accessed on Aug 23, 2021)} peer-reviewed citation database search engines to accomplish this.  The keywords are \textit{parking lot dataset}, \textit{parking lot images}, and \textit{parking lot database}. We refined the search to include only the works that propose datasets containing parking lot images. Finally, we reviewed only robust datasets, which must fulfill the following criteria.

First, the datasets' publicity is the most critical restriction considered in this work. Public data enable researchers to reproduce experiments, results and create their experiments without collecting and labeling novel datasets.
The cameras must be installed at fixed points since this is an expected feature in the real world (we do not consider, for instance, images collected via drone). 
The datasets must contain images collected on different days of the week, periods, and weather conditions to include the expected variability in a real scenario. 
Images collected from different camera angles are necessary to test classifiers' generalization power. For instance, a classifier is trained using images obtained from a camera angle different from the test images.
Finally, the ground truth is imperative to check and compare the results when using the datasets. 
Although restrictive, we consider the established criteria important for advancing the research about vision-based parking lot management.

As a result, three datasets were selected: the PKLot \citep{almeidaEtAl2013,almeidaEtAl2015}, the CNRPark-EXT \citep{amatoEtAl2016,amatoEtAl2017}, and the \ac{PLds} \citep{nietoEtAl2019}. 
Certain datasets despite being interesting, do not meet these restrictions, including the QuickSpotDB \citep{marmolSevillano2016}, CARPK \citep{liEtAl2019} and UAVDT \citep{duEtAl2018} datasets.

Following the selection of datasets, we proceeded further by surveying works that used these datasets to evaluate vision-based parking lot management approaches. Therefore, we used the Scopus and WoS search engines to identify cross-referenced works. Hence, we applied a snowballing approach \citep{WOHLIN:2014}, as the primary references of each dataset were used to find all the other works that cited each of them. These references are referred to as the primary references set. Further, the closure requirements, such as the objective, inclusion, and exclusion criteria, are as follows:

\begin{itemize}
\item Objective Criteria (OC)
\begin{itemize}
\item Document type: Articles
\item Availability:  Any
\end{itemize}
\item Inclusion Criteria (IC)
\begin{itemize}
\item IC{$_1$}: Include works that proposed the surveyed datasets.
\end{itemize}
\item Exclusion Criteria (EX)
\begin{itemize}
\item EX{$_1$}: Remove non-English works;
\item EX{$_2$}: Remove works that only mention the datasets as related work without using the datasets;
\item EX{$_3$}: Remove surveys or systematic reviews.
\end{itemize}
\end{itemize}

Based on the above rules, an initial number of 248 unique works were found. The IC{$_1$} included 5 \citep{almeidaEtAl2013,almeidaEtAl2015,amatoEtAl2016,amatoEtAl2017,nietoEtAl2019} works. The exclusion criteria  EX{$_1$}, EX{$_2$},  EX{$_3$} removed 10, 155, and 22 works, respectively. 
Thus, a total of \totalReviewedPapers\ works were identified for analysis.

\section{Parking Lot Datasets}\label{sec:datasets}

In this section, we bring forward an analysis on existing parking lot datasets available in the literature, i.e., PKLot, CNRPark, CNRPark-EXT, and PLds.

\subsection{PKLot Dataset}\label{subsec:PKLot}

The first version of the PKLot dataset, containing images of one parking lot and one camera angle, was released in \citet{almeidaEtAl2013}. The current version of the dataset was proposed by \citet{almeidaEtAl2015}, and contains 12,417 images collected from two different parking lots and three camera angles. The images were collected using a camera installed on the fourth and fifth floors of one building from the Federal University of Parana (UFPR) to generate the UFPR04 and UFPR05 subsets. Both UFPR04 and UFPR05 subsets contain images from the same parking lot yet collected under different camera angles and different days. The third subset of parking lot images was taken from the 10th floor of a building from the Pontifical Catholic University of Parana (PUCPR) and present a different camera angle and parking lot.

\begin{figure}[htbp]
    \centering
    \subfloat[UFPR04: Rainy]{
        \label{subfig:exemploUFPR04Rainy}
        \includegraphics[width=0.3\textwidth]{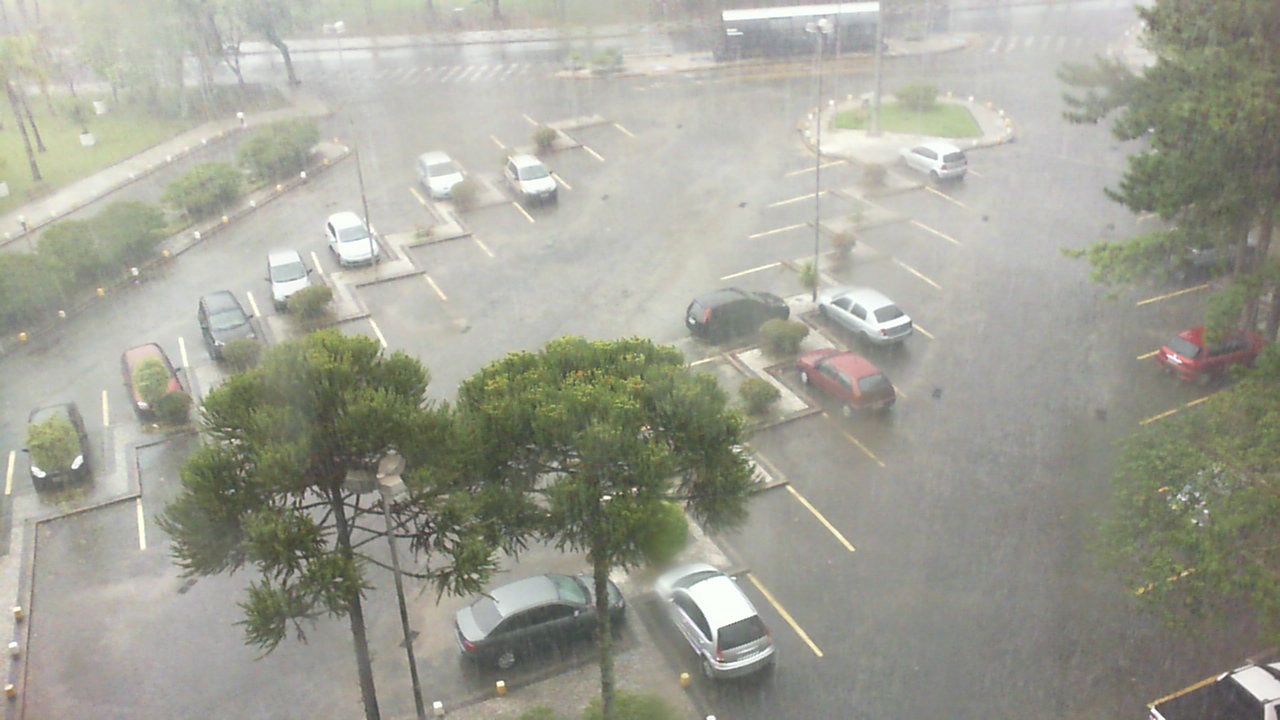}
    }
    \subfloat[UFPR05: Sunny]{
        \label{subfig:exemploUFPR05Sunny}
        \includegraphics[width=0.3\textwidth]{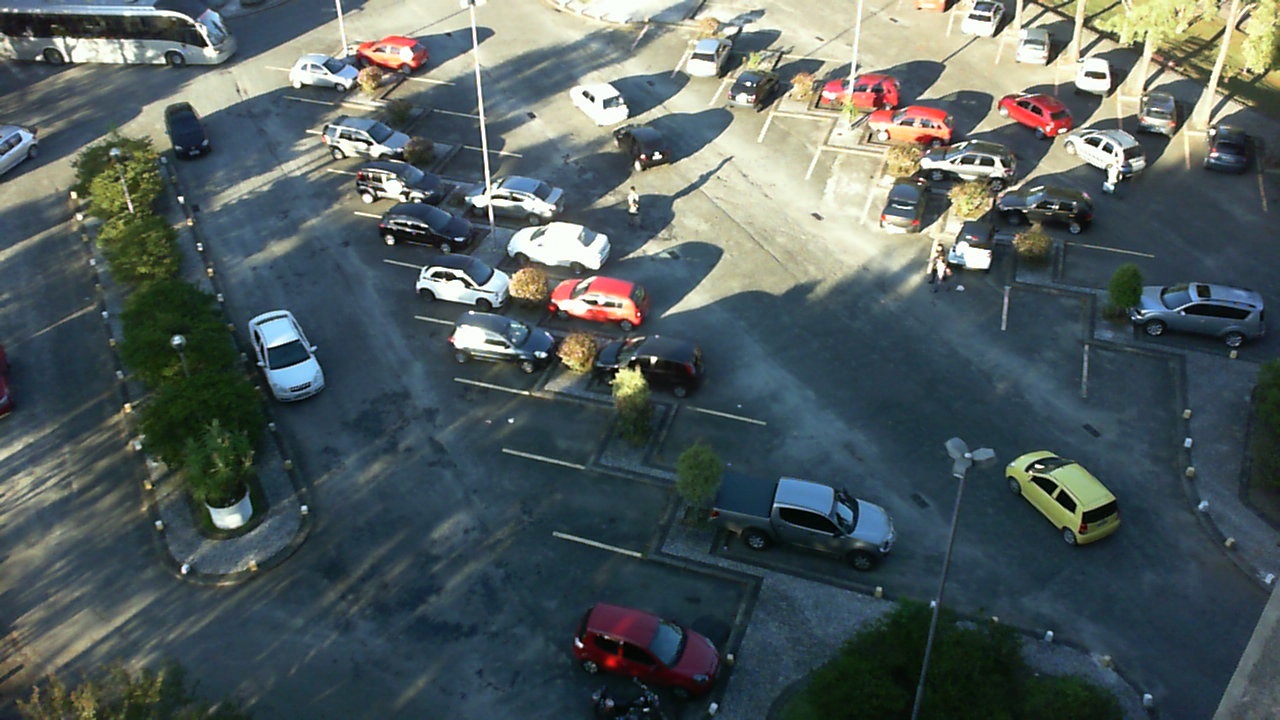}
    }
    \subfloat[PUCPR: Cloudy]{
        \label{subfig:exemploPUCPRCloudy}
        \includegraphics[width=0.3\textwidth]{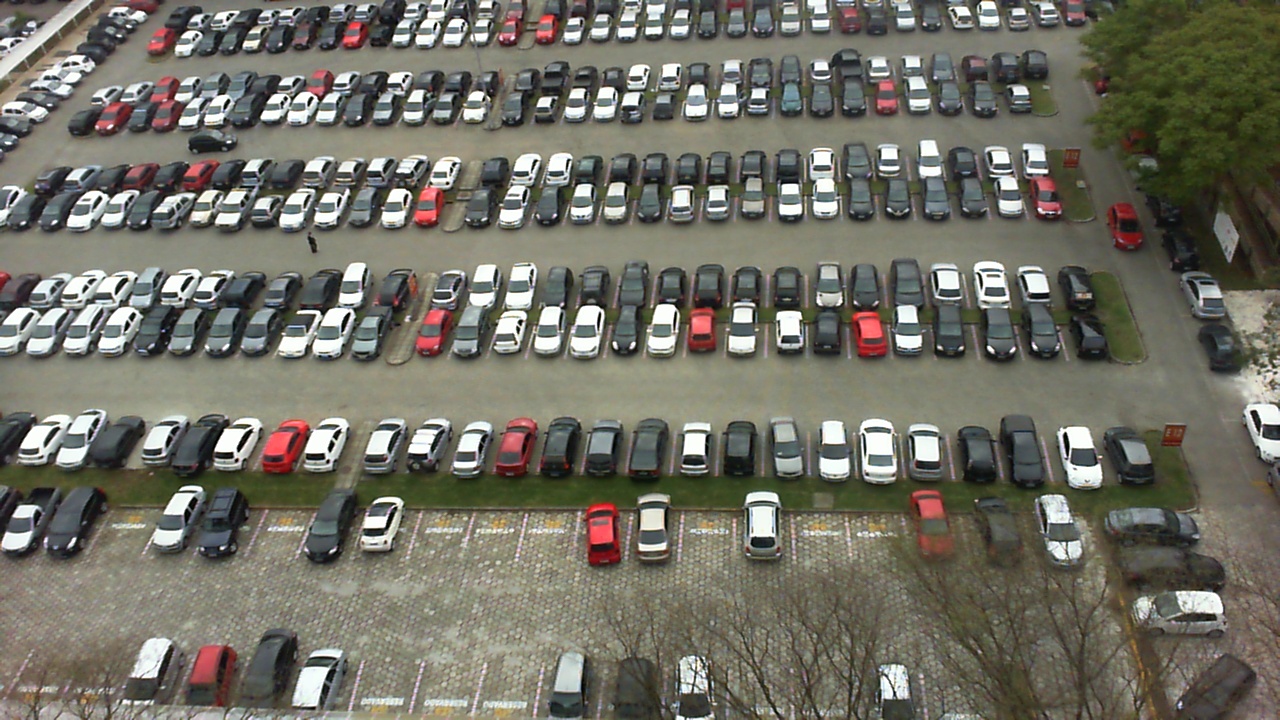}
    }
    \\
    \subfloat[UFPR05 with annotations]{
        \label{subfig:exemploAnotacoesUFPR05}
        \includegraphics[width=0.46\textwidth]{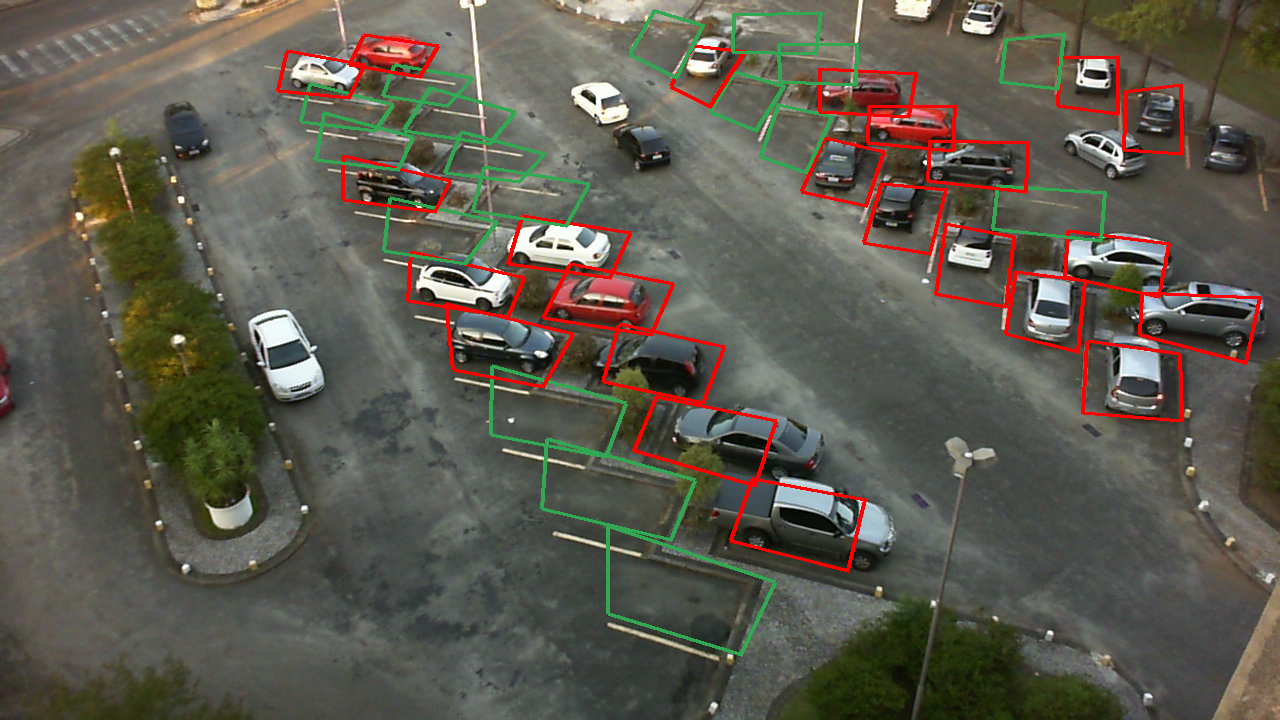}
    }
    \subfloat[PUCPR with annotations]{
        \label{subfig:exemploAnotacoesPUCPR}
        \includegraphics[width=0.46\textwidth]{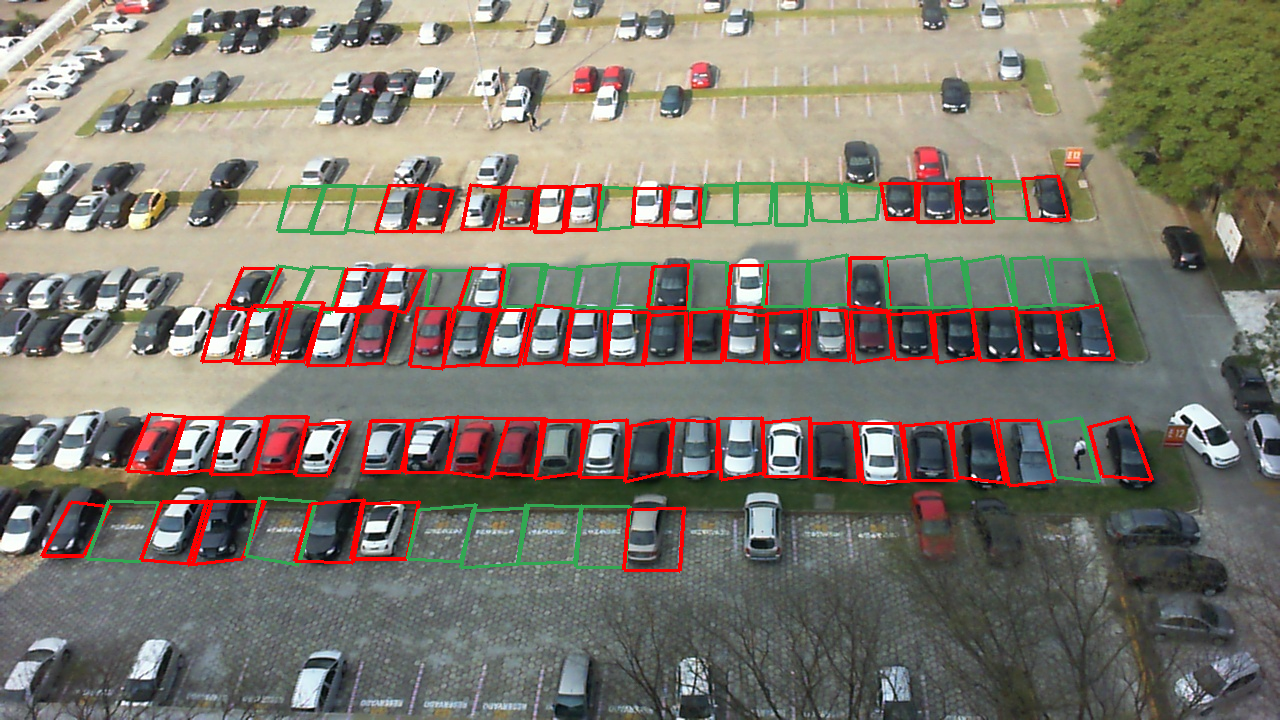}
    }
    \caption{PKLot image examples. Figures \protect\subref{subfig:exemploUFPR04Rainy}, \protect\subref{subfig:exemploUFPR05Sunny}, and \protect\subref{subfig:exemploPUCPRCloudy} show examples of different parking lots and weather conditions. Figures \protect\subref{subfig:exemploAnotacoesUFPR05} and \protect\subref{subfig:exemploAnotacoesPUCPR} show examples of images with the location and status (red for occupied and green for empty) of the parking spaces drawn.}
    \label{fig:ExamplesPKLot}
\end{figure}

The dataset contains 695,851 manually labeled individual parking spaces, such that 337,780 (48.6\%) are occupied, and 358,071 (51.4\%) are empty. Cropped images of the individual parking spaces are available altogether with the original dataset. 
All images are $1280 \times 720$ pixels in size and stored in the JPEG format. 
An \ac{XML} file containing four points of polygons representing each monitored parking space is available for each image. 
Additionally, a rotated rectangle representing the same information but with right angles (making it easier to crop the images) is available. 
Moreover, each parking space' status (empty/occupied) is also available in the \ac{XML} files.

Most of the parking spaces available in UFPR04 and UFPR05 parking lots are labeled in the dataset. 
In contrast, for the PUCPR, only 100 parking spaces were manually labeled. In comparison, approximately 300 parking spaces are visible to the human eye in such images. 
Images were collected under a 5 min interval during the daytime and labeled according to sunny, rainy, and cloudy weather. 
Figure \ref{fig:ExamplesPKLot} shows image examples from the PKLot dataset for different parking lots and weather conditions.

\subsection{CNRPark and CNRPark-EXT Datasets}\label{subsec:CNRPark-EXT}

\begin{figure}[ht]
    \centering
    \subfloat[Camera A]{
        \label{subfig:exemploCameraACNRPark}
        \centering
        \includegraphics[width=0.1\textwidth]{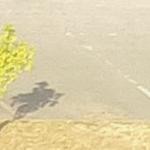}
        \includegraphics[width=0.105\textwidth]{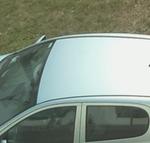}
        \includegraphics[width=0.1\textwidth]{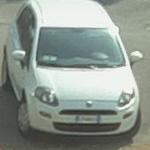}
    }\hspace{0.5cm}
    \subfloat[Camera B]{
        \label{subfig:exemploCameraBCNRPark}
        \centering
        \includegraphics[width=0.1\textwidth]{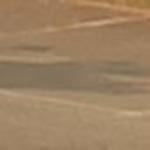}
        \includegraphics[width=0.1\textwidth]{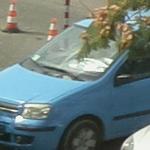}
        \includegraphics[width=0.1\textwidth]{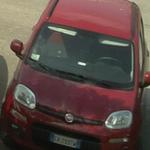}
    }
    \caption{CNRPark image examples.}
    \label{fig:ExamplesCNRPark}
\end{figure}

The CNRPark dataset, proposed in \citet{amatoEtAl2016}, contains images collected from a single parking lot using two different camera angles. 
Images were taken considering a 5 min interval during daytime in sunny conditions on two different days for each camera angle. 
The dataset contains 12,584 parking spaces, where 4,181 (33.2\%) are free, and 8,403 (66.8\%) are occupied. 
Only the segmented parking spaces are publicly available, where all images were resized to $150 \times 150$ pixels. 
In addition to certain parking spaces being angled relative to the camera, the images were segmented using non-rotated squares, which may have resulted in the inclusion of undesired areas or exclusion of certain regions of the parking spaces or cars. Examples of the CNRPark dataset can be seen in Figure \ref{fig:ExamplesCNRPark}.

\begin{figure*}[ht]
    \centering
    \subfloat[Camera 3: Rainy]{
    \label{subfig:exemploCNRParkRainy}
        \centering
        \includegraphics[trim={0 0 0 2.3cm}, clip, width=0.26\textwidth]{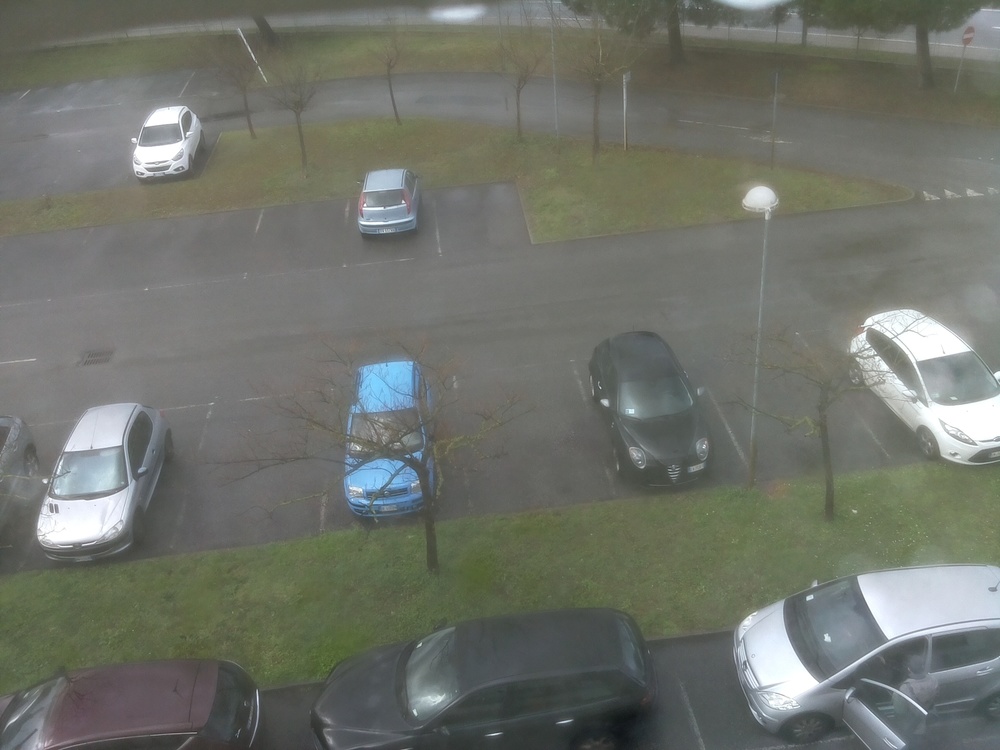}
    }
    \subfloat[Camera 5: Sunny]{
        \label{subfig:exemploCNRParkSunny}
        \centering
        \includegraphics[trim={0 0 0 2.3cm}, clip, width=0.26\textwidth]{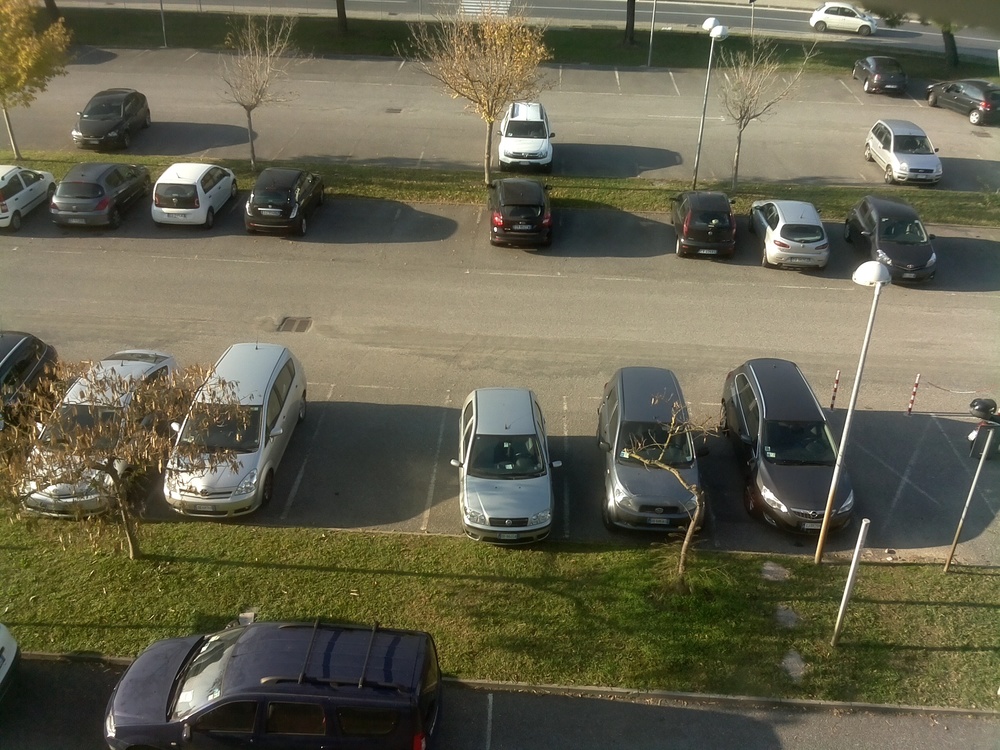}
    }
    \subfloat[Camera 9: Overcast]{
        \label{subfig:exemploCNRParkOvercast}
        \centering
        \includegraphics[trim={0 0 0 2.3cm}, clip, width=0.26\textwidth]{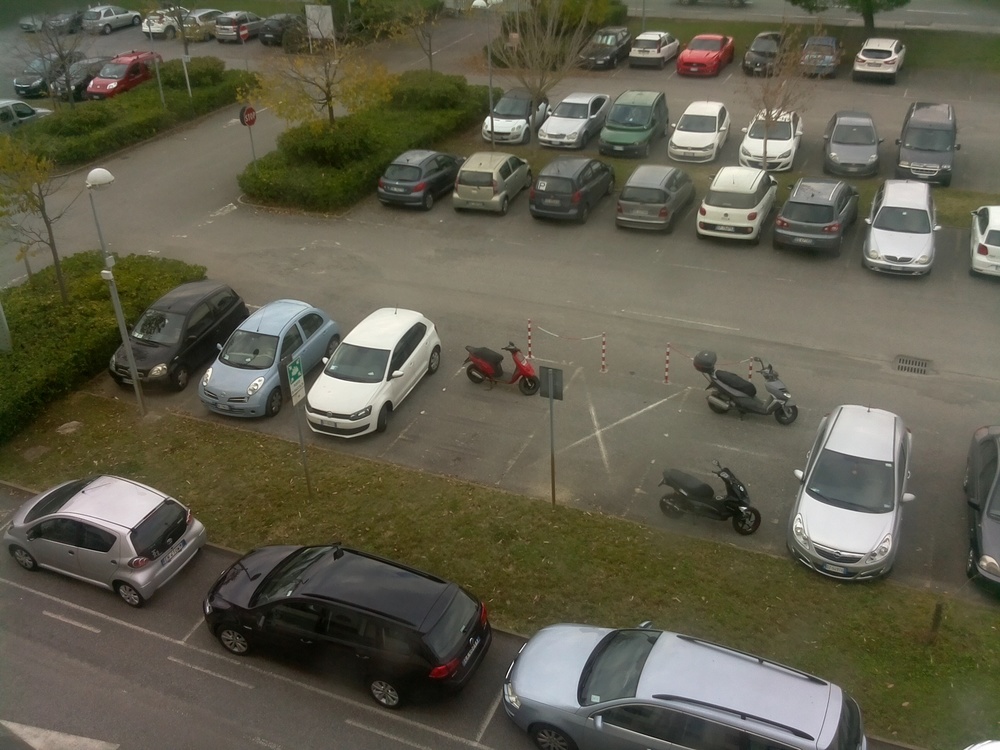}
    }\\
    \subfloat[Camera 1 with annotations]{
        \label{subfig:exemploAnotacoesCNRParkC1}
        \centering
        \includegraphics[trim={0 0 0 2.3cm}, clip, width=0.4\textwidth]{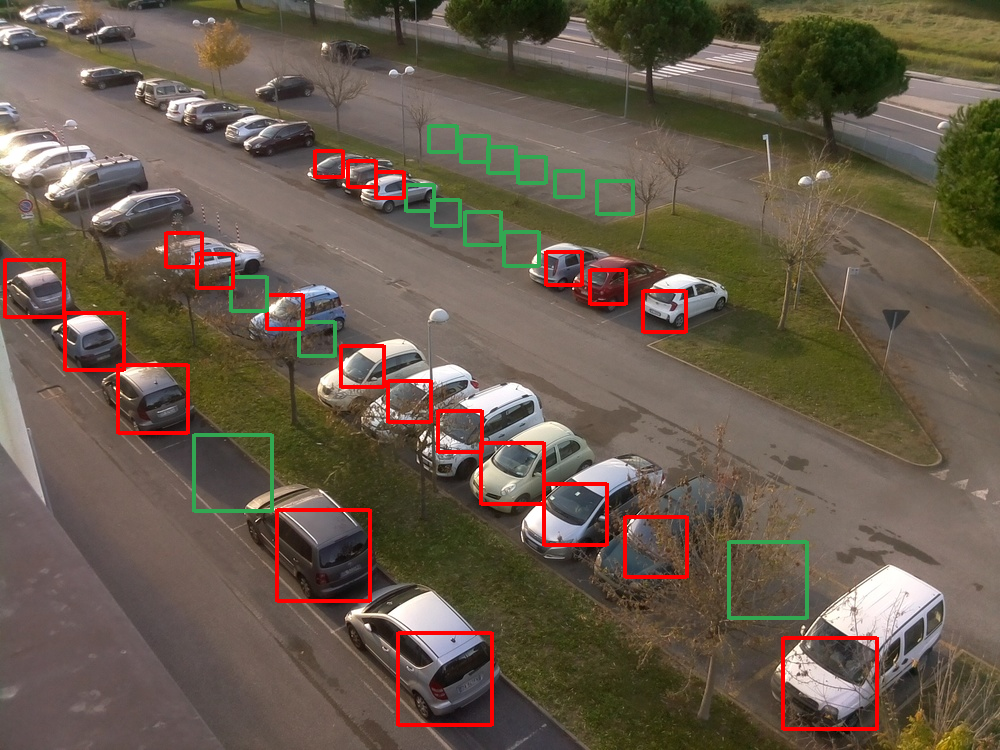}
    }
    \subfloat[Camera 2 with annotations]{
        \label{subfig:exemploAnotacoesCNRParkC2}
        \centering
        \includegraphics[trim={0 0 0 2.3cm}, clip, width=0.4\textwidth]{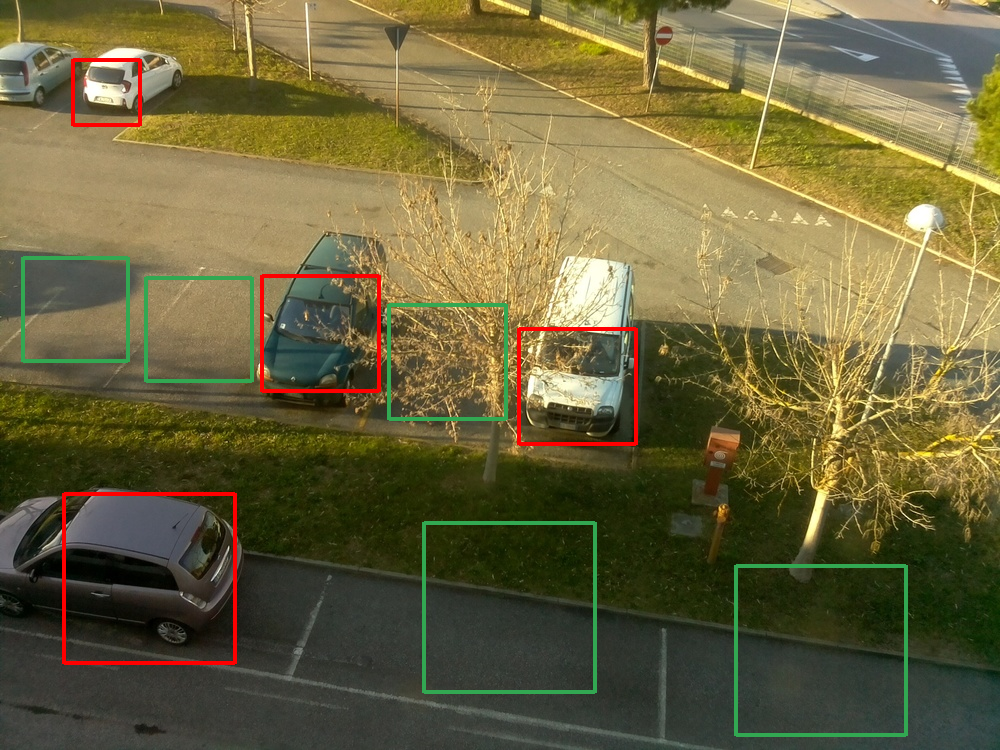}
    }
    \caption{CNRPark-EXT image examples. Figures \protect\subref{subfig:exemploCNRParkRainy}, \protect\subref{subfig:exemploCNRParkSunny}, and \protect\subref{subfig:exemploCNRParkOvercast} show examples of different parking lots and weather conditions. Figures \protect\subref{subfig:exemploAnotacoesCNRParkC1} and \protect\subref{subfig:exemploAnotacoesCNRParkC2} show examples of images with the location and status (red for occupied and green for empty) of the parking spaces drawn.}
    \label{fig:ExamplesCNRParkExt}
\end{figure*}

The CNRPark was extended to create the CNRPark-EXT dataset in \citet{amatoEtAl2017}. 
The authors used nine cameras to capture images from a single parking lot at different angles. Images were acquired at a 30 min interval.
The cameras are not synchronized with each other. 
Similar to the PKLot dataset, the authors separated the images according to the weather.
The images are $1000 \times 750$ pixels in size, with a total of 4,278 JPEG parking lot images available.
The dataset also encompasses \ac{CSV} files that indicate the monitored parking spaces' coordinates, represented by non-rotated squares (see Figures \ref{subfig:exemploAnotacoesCNRParkC1} and \ref{subfig:exemploAnotacoesCNRParkC2}).
Cropped images of the individual parking spaces are also available, where the images were resized to $150 \times 150$ pixels. 
When considering the images of the original CNRPark altogether with the CNRPark-EXT datasets, there are 157,549 labeled parking spaces available, where 69,839 (44,3\%) are free, and 87,710 (55,7\%) are occupied.

The CNRPark-EXT is considered an extension of the original CNRPark dataset by the authors, and only \citet{amatoEtAl2016} used the CNRPark images without the CNRPark-EXT dataset. 
Thus, we have considered CNRPark- EXT concatenated with the original CNRPark as a single dataset. Figure \ref{fig:ExamplesCNRParkExt} shows certain examples of the images contained in the CNRPark-EXT dataset. They were acquired from different camera angles and under different weather conditions, and have been presented along with certain annotated image examples.

\subsection{PLds Dataset}\label{subsec:plds}

The \acf{PLds} was proposed in \citet{nietoEtAl2019}, and contains $1280 \times 960$ pixel images collected from three different camera angles at the Pittsburgh International Airport parking lot.
Images were taken under several time intervals, ranging from a few seconds to several minutes, i.e., 15 s to 30 min\footnote{We considered the time tag available in the top left corner of the images.}. 
Several climate conditions are available in the dataset, including snow and rain. Light conditions also include nighttime images.

\begin{figure}[ht]
    \centering
    \subfloat[Rainy]{
        \label{subfig:exemploPLdsRainy}
        \includegraphics[trim={0 0 9.0cm 4.0cm}, clip, width=0.29\textwidth]{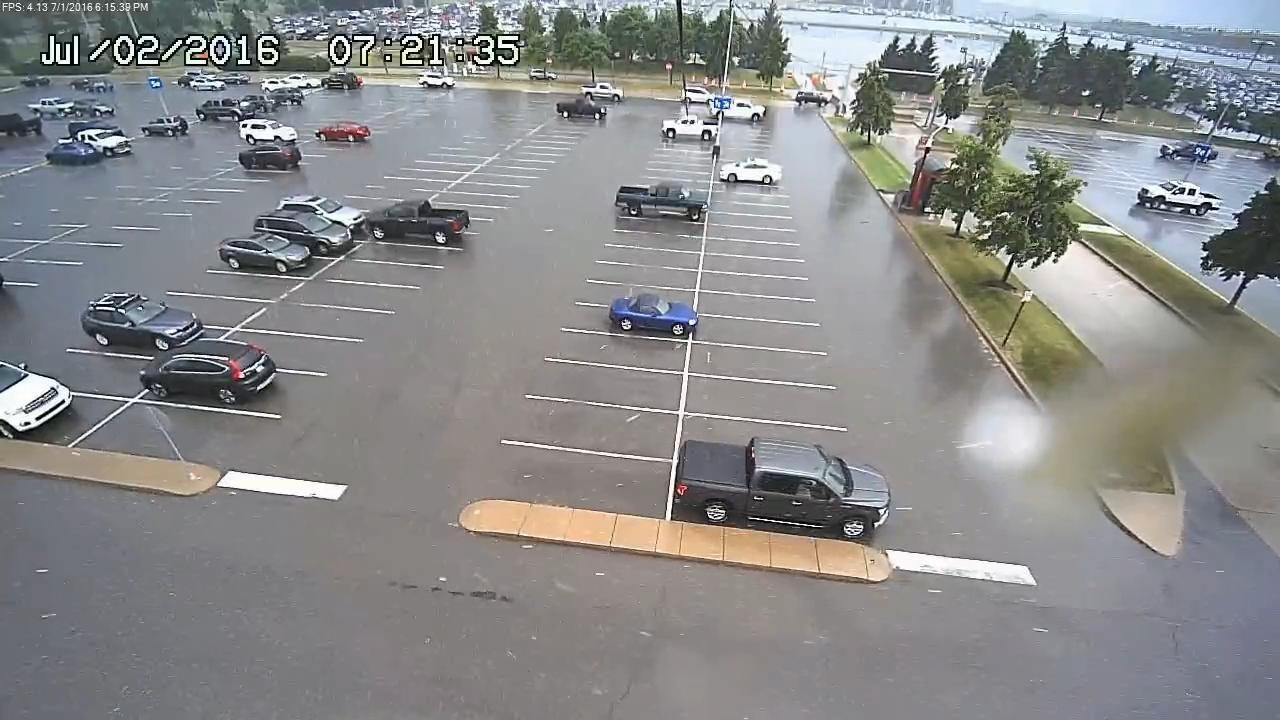}
    }
    \subfloat[Sunny]{
        \label{subfig:exemploPLdsSunny}
        \centering
        \includegraphics[trim={9.0cm 0 0 4.0cm}, clip, width=0.29\textwidth]{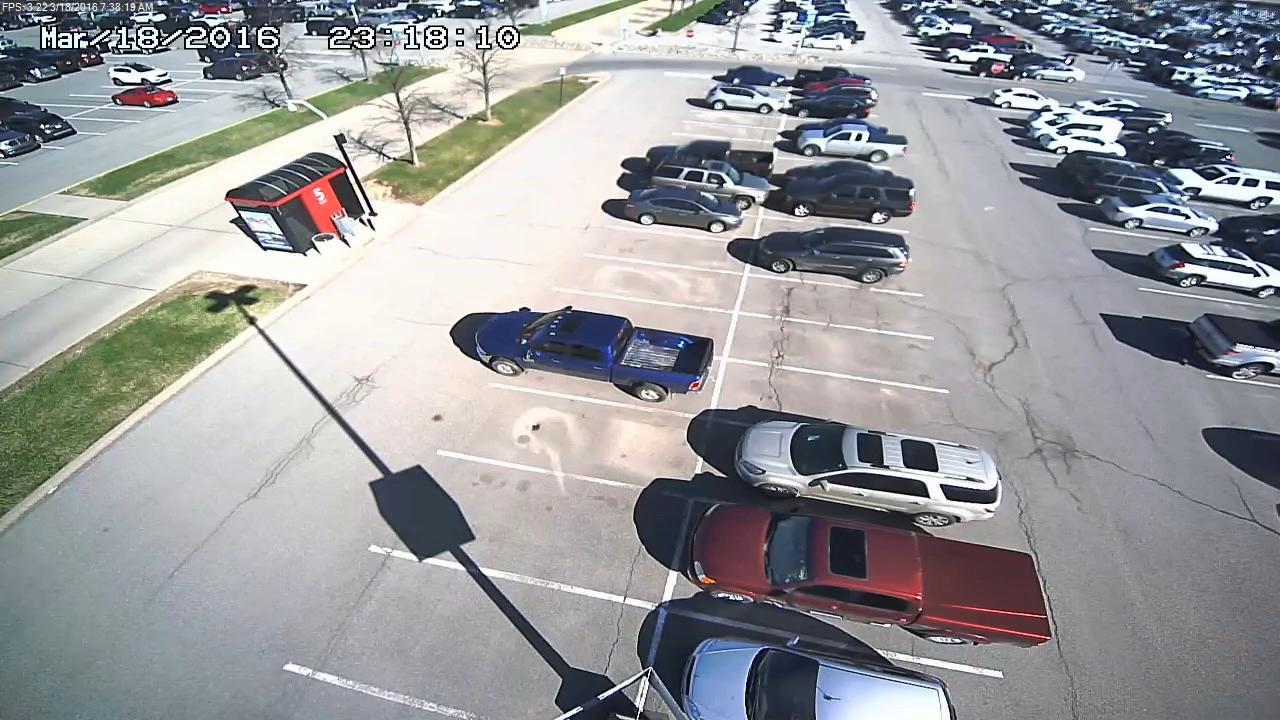}
    }
    \subfloat[Snow]{
        \label{subfig:exemploPLdsSnow}
        \includegraphics[trim={0 0 9.0cm 4.0cm}, clip, width=0.29\textwidth]{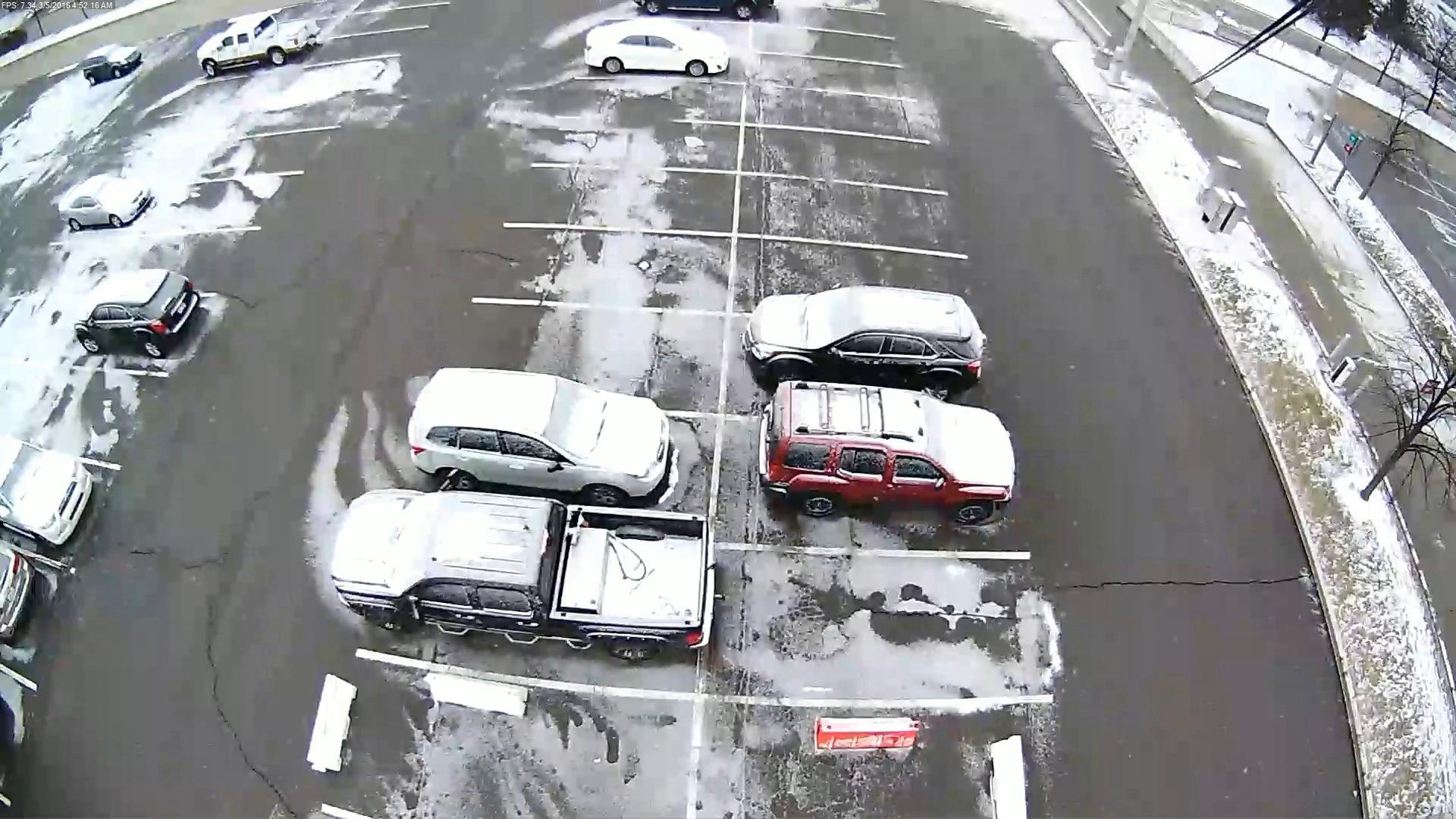}
    }
    \\
    \subfloat[PLds image with annotations]{
        \label{subfig:exemploAnotacoesPlds1}
        \includegraphics[width=0.44\textwidth]{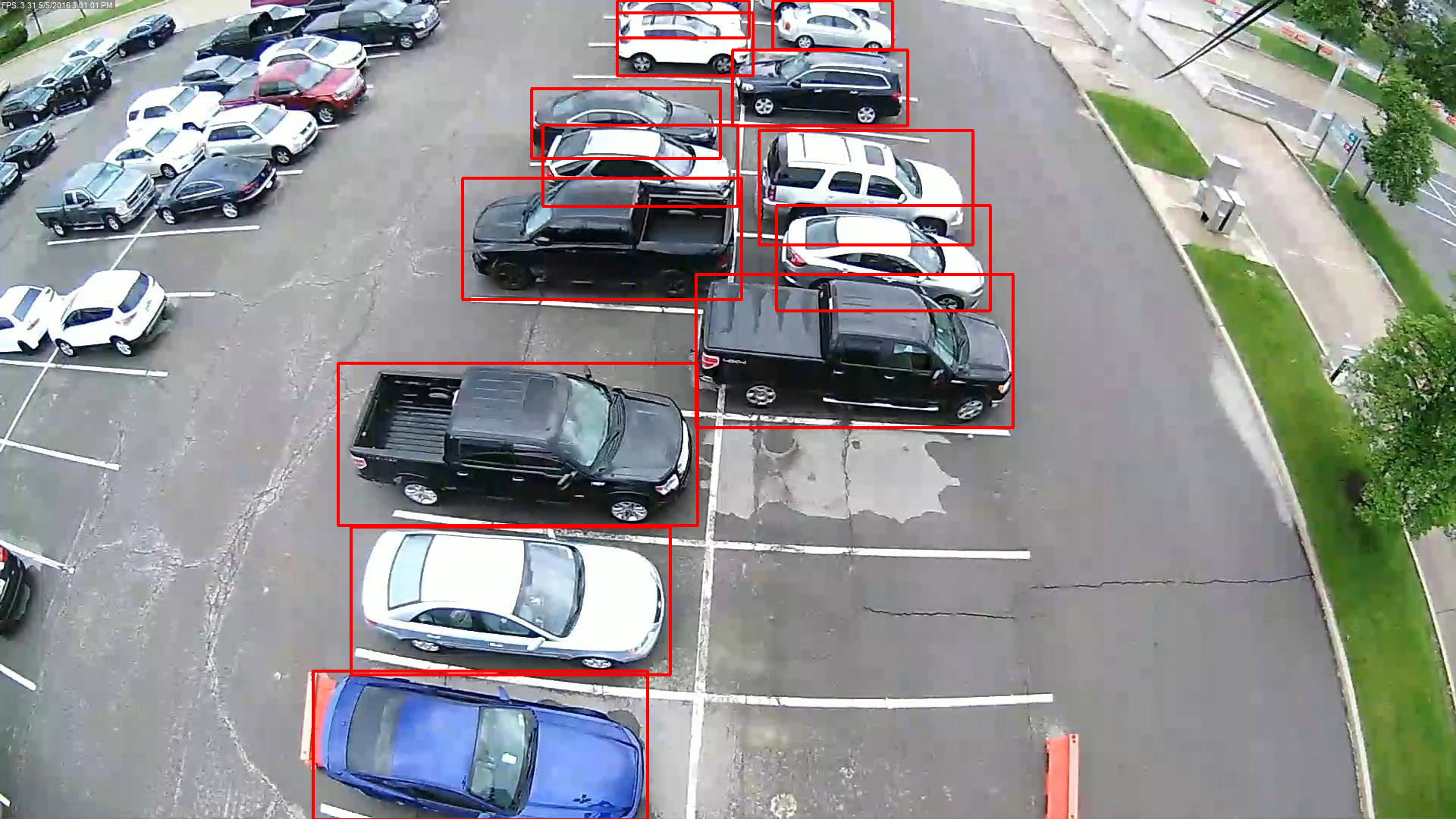}
    }
    \subfloat[PLds image with annotations]{
        \label{subfig:exemploAnotacoesPlds2}
        \includegraphics[width=0.44\textwidth]{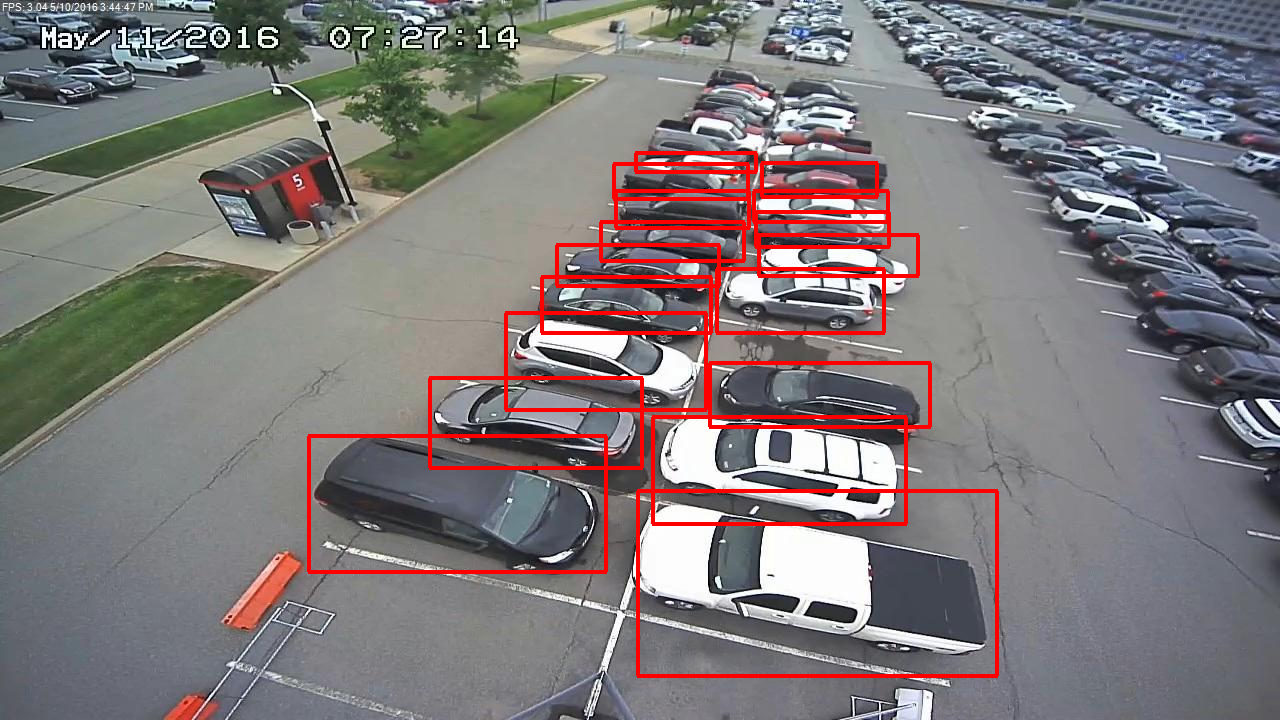}
    }
    \caption{PLds image examples. Figures \protect\subref{subfig:exemploPLdsRainy}, \protect\subref{subfig:exemploPLdsSunny}, and \protect\subref{subfig:exemploPLdsSnow} shows examples of different weather conditions and camera angles. Figures \protect\subref{subfig:exemploAnotacoesPlds1} and \protect\subref{subfig:exemploAnotacoesPlds2} show examples of images with cars annotations.}
    \label{fig:ExamplesPLds}
\end{figure}

The images are stored in JPEG format and the annotations for each image has been provided by the authors in XML files. In addition, non-rotated bounding boxes for parked cars are provided (annotations for empty parking spots are not available). An appealing feature of this dataset is that a subset of PLds containing 100 images is synchronized between two different camera angles. Image examples of the PLds dataset, including different weather conditions and annotated images, are shown in Figure \ref{fig:ExamplesPLds}.

The dataset contains 8,340 images\footnote{We did not consider the sequence of images isshk\_1955 to isshk\_2146, and isshk\_2598 to isshk\_2681 since the images seem to repeat.}. 
In the original dataset, the images are not labeled according to the climate nor to luminosity conditions. 
As a byproduct contribution of this review, we manually classified the images between day/night and climate conditions (see the summary of the images after this classification in Section \ref{subsec:datasetsSummary}). 
We made the list of the images classified according to climate and luminosity publicly available at \url{github.com/paulorla/datasets/tree/main/PLds}.

\section{State of the Art Review}\label{sec:stateoftheartreview}

In this section, we present the review of the works regarding computer vision-based approaches for parking lot management. Specifically, we divide our analysis and discussion based on the different approaches regarding parking lot management. 
In practice, we first discuss methods for the individual parking spaces classification (Section \ref{sec:indivitualParkingSpots}), followed by approaches for the automatic parking space detection (Section \ref{sec:parkingSpacesSegmentation}), and finally, methods for car detection and counting (Section \ref{sec:carsCounting}).

\subsection{Individual parking spaces classification}
\label{sec:indivitualParkingSpots}

\begin{figure}[ht]
    \centering
    \subfloat[Vacant]{
        \label{subfig:exampleVacant}
        \centering
        \includegraphics[width=0.12\textwidth]{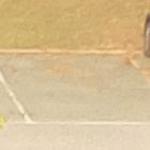}
    }\hspace{0.5cm}
    \subfloat[Occupied]{
        \label{subfig:exampleOccupied}
        \centering
        \includegraphics[width=0.12\textwidth]{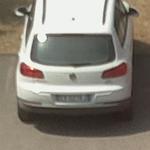}
    }
    \caption{Individual parking spaces classification example (images from CNRPark-EXT).}
    \label{fig:parkSporClassificationExample}
\end{figure}

The individual parking spot classification task can be modeled as the binary problem of classifying individual parking spaces based on whether occupied or vacant.  
This problem is exemplified in Figure \ref{fig:parkSporClassificationExample}. Each image containing a simple parking spot is fed to a classifier to define its status as vacant, as exemplified in Figure \ref{subfig:exampleVacant}, or occupied, as in Figure \ref{subfig:exampleOccupied}.
The methods were split into two major groups: methods based on Feature Extraction (Section \ref{subsec:featureExtractionMethods}) and \ac{DL} methods (Section \ref{subsec:deepLearningMethors}).

\subsubsection{Feature Extraction-based Methods}
\label{subsec:featureExtractionMethods}

\begin{figure}[ht]
    \centering
    \resizebox{\textwidth}{!}{
        
    \tikzsetnextfilename{./tikz/featureExtrMethodsDiag}%
    \input{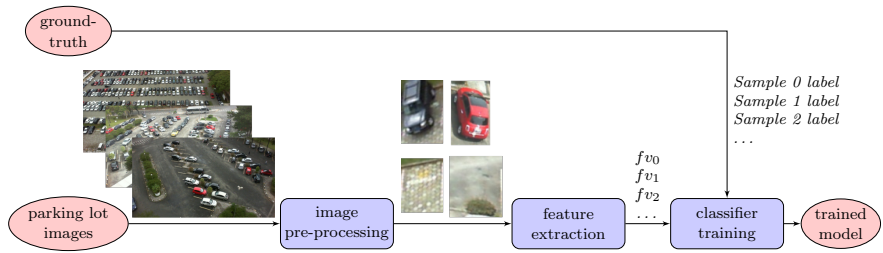}%

    }
    \caption{Feature Extraction based methods high-level scheme. In the illustration, $fv_i$ refers to the feature vector extracted from the sample (parking space image) $i$.}
    \label{subfig:FeatureExtractionDiagram}
\end{figure}

The training phase of the methods based on feature extraction follows the scheme depicted in Figure \ref{subfig:FeatureExtractionDiagram}.
As an \textit{input image}, we consider the entire image of the parking lot, acquired via a camera.
The \textit{image pre-processing} step is used to segment the complete image in individual parking spaces.
Some authors may also use techniques to make the image more suitable for feature extraction during the \textit{image pre-processing} step, such as image scaling and histogram equalization.
One or more feature vectors may be extracted from the images in the \textit{feature extraction step}, such as the \ac{LPQ} \citep{ojansivuHeikkila2008}, the \ac{LBP} \citep{ojalaPietikainen1999}, or the \ac{HOG} \citep{dalalTriggs2005}.

A classifier, such as an \ac{SVM} or \ac{MLP}, is then used in the \textit{model training} step. The feature vectors extracted from the images, and the ground-truth of each image (indicating the correct label of each parking space), are fed to train the classifier during this step.
Thus, this final step results in the creation of the trained model, which can be used to classify unseen images.

The use of texture-based features, such as \ac{LBP}, \ac{LPQ}, and \ac{QLRBP} \citep{LanZhouTang2016}, is common when dealing with the individual parking spaces classification problem. In
\citet{almeidaEtAl2013} is proposed the use of \ac{LPQ} and \ac{LBP} textures as feature vectors and \acp{SVM} as classifiers.
In this work, the first version of the PKLot dataset was introduced.
The work and the dataset were extended in \citet{almeidaEtAl2015}, where the entire PKLot dataset was released. Ensembles of \acp{SVM} trained using \ac{LPQ}/\ac{LBP} as features were used for classification.

Owing to the possible changes caused by luminosity, camera shifts, and parking lot area changes, the authors in \citet{almeidaEtAl2018,almeidaElAl2020} considered the individual parking spaces classification from the perspective of a concept drift. Therefore, the authors in \citet{almeidaEtAl2018} considered their custom framework for dealing with concept drifts (called Dynse) and employed \ac{LBP} features. For the experiments, the PKLot dataset was used. 
The parking lot images are presented day by day in an ordered fashion. Therefore, all current-day instances must be classified according to 100 randomly sampled instances (images) from the previous day used for training.
In \citet{almeidaElAl2020}, the authors assessed several datasets, including the PKLot, to search for concept drifts' evidence. Using the same features and data split of \citet{almeidaEtAl2018}, the authors showed that a static or naïve classifier yielded results that are worse than approaches tailored to address concept drifts in the PKLot dataset.

In \citet{suwignyoSetyawanYohanes2018} the \ac{QLRBP} was employed as texture features from the color images of the parking spaces. 
The authors use \ac{k-NN} and \acp{SVM} as classifiers. 
For the tests, 6,000 individual parking spaces of the UFPR04 subset were used.
\citet{hammoudiEtAl2018,hammoudiEtAl2018b,hammoudiEtAl2019,hammoudiEtAl2020} also proposed the use of \ac{LBP}-based features to classify parking lot images. 
In these four quite similar works, the authors used a \ac{k-NN} classifier and small image subsets (3,000 to 6,000 segmented images) of the PKLot for the tests. 
However, the manner in which the authors grouped the images into subsets is not clear. Further, the authors in \citet{hammoudiEtAl2019} also included \ac{SVM} classifiers for the tests and tested the changes within parking lots employing a subset of the PKLot and CNRPark-EXT to conduct the tests.

In \citet{dizon2017development} \ac{LBP} and \ac{HOG} were used as features descriptors for a linear \ac{SVM} classifier.
The authors also employed a background subtraction approach with \ac{AMF}.
The results reported in this study indicated that a classifier trained using only the \ac{HOG} exhibited good results in the UFPR04 subset from PKLot.
\citet{thike2019parking} used \ac{ULBP} in a complemented image combined with \ac{MSE}, which is used to classify a parking space based on a threshold applied to the \ac{MSE} output. 
The authors tested 1,000 images from the PKLot dataset from different weathers (it is unclear how images were selected).

In the work of \citet{dornaikaEtAl2019}, \ac{SVM} and \ac{k-NN} classifiers are trained with textural features extracted from different scales of the images.
The authors used subsets of the PKLot and CNRPark datasets. 
In addition, a custom-built dataset, including images from the CNRPark and ImageNet \citep{dengEtAl2009Imagenet} was used in the tests.
\citet{irfanEtAl2020} proposed Gray-Level Co-Occurrence Matrixes (\acs{GLCM}) as texture features. 
The test images are classified as occupied or empty according to their similarity to the train images. 
The authors used only 60 images from the PUCPR subset of the PKLot for the tests.
A combination of color and texture features is employed in \citet{magoKumar2020}. In addition, the authors put to the test Neural Networks, \acp{SVM}, \acp{k-NN}, and Naïve Bayes classifiers. 
The authors employed the PKLot for the tests, but no details about the testing procedure were given.

The use of the pixel values under different color spaces is another approach commonly employed by authors, such as \citet{baroffioEtAl2015,AhrnbomAstromNilsson2016,hadiGeorge2019}.
\citet{baroffioEtAl2015} compute the histograms in \ac{HSV} color space directly in smart cameras. 
The histograms are sent to a central, which uses it as features for an \ac{SVM} classifier with a linear kernel (also seen in \citet{bondi2015ez}, from the same authors).
The PKLot dataset was used in the tests. 
The authors claim that energy and bandwidth can be saved by pre-processing the images inside the cameras and sending only the feature vectors or compressed images to a central. 
\citet{AhrnbomAstromNilsson2016} tested \ac{SVM} and \ac{LR} classifiers trained using feature channels, i.e., the individual color channels of an image. 
Tests were conducted using the PKLot dataset.

After pre-processing the images using \ac{DWT}, grayscale conversion, and binary thresholding, the authors in \citet{vitekMelnicuk2018} employed a simple image average as their feature descriptor. A threshold is applied in the feature descriptor for classification. 
PKLot and CNRPark-EXT datasets were used for the tests. 
In \citet{hadiGeorge2019}, the chromatic gradient analysis of the images is used to classify the individual parking spaces. 
In addition, the authors proposed an adaptive weather analysis technique to improve the results. Moreover, the classification of parking was based on a threshold, and the PUCPR subset of the PKLot was used to conduct the tests. However, details regarding the test procedure and quantitative results are absent.

\citet{amatoEtAl2018} proposed an approach based on background modeling and the Canny edge detector.
The approach was developed to be deployed on smart cameras. The CNRPark-EXT dataset was used during the tests.
\citet{raj2019vacant} also employed the Canny Edge detector but combined with a transformation to the LUV color space to generate the features.
A random forest classifier is used for classification.

Bag of features representations of the features is employed in \citet{vargheseSreelekha2019} and \citet{moraEtAl2018}.  
\citet{vargheseSreelekha2019} proposed an approach that uses an \ac{SVM} classifier and a bag of features representation. 
This representation combines the \ac{SURF} \citep{bayEtAl2008} descriptor and color features to classify the individual parking spaces of the PKLot and CNRPark-EXT datasets.
\citet{moraEtAl2018} also proposed an approach using a bag of features for classifying the individual parking spaces.
The authors use the SIFT \citep{Lowe1999} algorithm to extract the features and use an \ac{SVM} with a radial basis kernel as a classifier.
The authors evaluated their method with the PKLot, considering camera angle, climate, and parking lot changes.
The authors also propose a \ac{DL}-based approach, further described in Section \ref{subsec:deepLearningMethors}.
 
The \ac{HOG} descriptor's used as features and the \ac{SVM} classifier is explored by \citet{bohushEtAl2018} and by \citet{vitekMelnicuk2018}.
In \citet{bohushEtAl2018}, a subset of the PKLot dataset containing 2,135 images is used for the tests. 
However, the authors do not make clear how the images were selected. 
The authors also propose a segmentation method based on classical image processing methods, further described in Section \ref{sec:parkingSpacesSegmentation}. 
In \citet{vitekMelnicuk2018}, a lightweight approach developed to be deployed on Smart Cameras is proposed. 
Information about the parking angle is concatenated with the \ac{HOG} feature vector to improve the results. 
Tests were made using the PKLot and two private datasets.

\subsubsection{Deep Learning Based Methods}\label{subsec:deepLearningMethors}

\begin{figure*}[ht]
    \centering
    \resizebox{\textwidth}{!}{
        
    \tikzsetnextfilename{./tikz/deepLearningMethodsDiagram}%
    \input{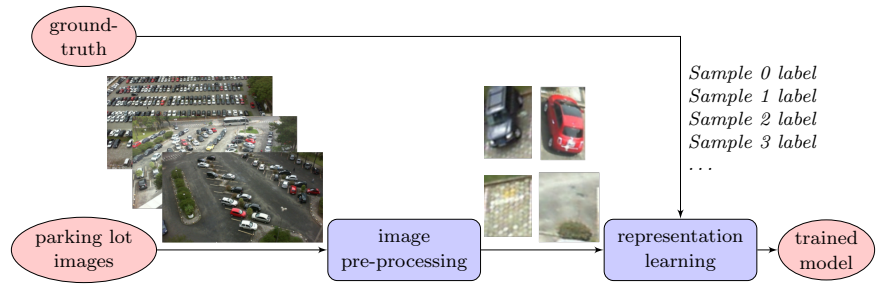}%

    }
    \caption{\ac{DL}-based methods high level scheme.}
    \label{subfig:DeepLearningMethodsDiagram}
\end{figure*}

\ac{DL}-based methods follow a workflow similar to feature-based approaches, except with the feature extraction and classifier training parts being joined together in a \textit{representation learning} block.
Here, feature engineering is not employed since \ac{DL} models aim to learn the representation of the parking spaces.
Also, the classifier is generally a part of the \ac{DL} model.
This workflow can be seen in Figure \ref{subfig:DeepLearningMethodsDiagram}.

The \textit{image-processing} step follows the same concept described for feature-based methods (Section \ref{subsec:featureExtractionMethods}).
It may also include data augmentation approaches, aiming to increase the number of samples and their variability.
The \textit{representation learning} in this problem can be divided into (i) transfer learning of well-known convolutional networks for classification, such as LeNet \citep{lecunEtAl1998} and AlexNet \citep{krizhevskyEtAl2012}; (ii) the proposal of a custom convolutional model, generally based on these well-known networks; and (iii) the use of \ac{DL} networks for object detection or segmentation, such as Faster-RCNN \citep{renEtAl2015} and Mask R-CNN \citep{kaimingEtAl2017}.
It is also noteworthy that in \citep{AcharyaYanKhoshelham2018}, authors used the SVM classifier to replace the softmax function of neural networks.

Transfer learning \citep{yosinski2014transferable}, where a network is first trained in a generic dataset and then fine-tuned in a parking lot dataset, is a common approach. 
LeNet and AlexNet networks are popular due to their compactness. 
Both are used in \citet{nyambalKlein2017} to classify parking lots. 
The authors use the PKLot and a private dataset, although it is unclear how the tests were performed for the PKLot. 
The usage of AlexNet is also seen in \citet{diMauroEtAl2016}. 
It focuses on optimizing a model with only a few samples, either PKLot or a private dataset.

The authors in \citet{dingEtAl2019} proposed to add residual blocks in the Yolov3 \citep{RedmonFarhadi2018} to extract more granular features. The modified network is used to classify the parking lot's images. 
Vehicle images from the PASCAL VOC \citep{everingham2010pascal} and COCO \citep{lin2014microsoft} datasets are used to train this network. 
Then, it is fine-tuned using some images of the PUCPR subset of the PKLot. 
Images of PUCPR were also used during the tests. 
The authors do not clarify how the images were split between the training and testing sets.

Many works proposed lightweight models based on well-known convolutional networks, such as LeNet, AlexNet, and VGGNet \citep{simonyanZisserman2014}. 
These custom models are primarily convolutional networks similar to the original networks but with fewer layers, i.e., shallow networks.
These models are usually developed for low-power and restricted processing capabilities devices, such as smart cameras.
Works such as \citet{amatoEtAl2016, amatoEtAl2017, polprasertEtAl2019, valipourEtAl2016, AcharyaYanKhoshelham2018, buraEtAl2018, MerzougEtAl2019, rahman2020convolutional, KolharAlameen2021} can be grouped in these lightweight versions, where most authors use the PKLot dataset for the tests.
Private datasets were included in \citet{AcharyaYanKhoshelham2018, polprasertEtAl2019,MerzougEtAl2019, buraEtAl2018}.

The authors in \citet{amatoEtAl2016} proposed the mAlexNet, based on the AlexNet network, and executed experiments in CNRPark. 
\citet{amatoEtAl2017} used the extended version of the dataset, the CNRPark-EXT. 
Their mAlexNet can cope with the parking lot and camera angle variations with tiny accuracy drops in many scenarios (the authors also included some tests in the PKLot). 
Similarly, \citet{amatoEtAl2018} employed the mAlexNet to classify the parking spaces using smart-cameras and used the CNRPark-EXT dataset for the tests.
\citet{rahman2020convolutional} also employed mAlexNet but changed the kernel size of the first layer. No significant difference in the final results was found.
\citet{nguyen2021adaptive} evaluated the mAlexNet, AlexNet, and  MobileNet \citep{howard2017mobilenets} networks with a width multiplier of 0.5 (which presented better performance) using Camera A of the CNRPark and a private dataset. They aimed an approach for low-cost hardware, but the processing cost (and time) after image capture was left unclear.

\citet{buraEtAl2018} also proposed a lightweight version of the AlexNet network in their work, wherein they used subsets of the PKLot, CNRPark-EXT, and a private dataset to conduct the tests. Nevertheless, information regarding the availability of the private dataset and the manner in which the images were selected for the subsets is absent.

A lighter version of the AlexNet network is also proposed in \citet{AliMohamed2021}. The authors removed one of the convolutional layers of the original network, together with some minor modifications. The PKLot dataset was used for the tests.

The authors in \citet{valipourEtAl2016} trained a VGGNet-F \citep{simonyanZisserman2014} model with the ImageNet \citep{dengEtAl2009Imagenet} dataset and fine-tuned it with PKLot images, which resulted in better generalization capabilities (considering camera and parking lot changes) when compared with the feature extraction methods used in \citet{almeidaEtAl2015}. 
\citet{AcharyaYanKhoshelham2018} proposed an approach wherein a VGGNet-F model is trained for feature extraction and inputs features to an \ac{SVM} classifier. 
The authors employed 5-fold cross-validation in the PKLot images during the test phase. \citet{ZhangEtAl2019} proposed a modified version of the VGG16 network and a custom network to classify whether the parking spaces are empty or occupied.
The authors also proposed applying image transformations to get a top view of the parking lots and employed the PKlot dataset during the tests.

The original VGG16 network is used by \citet{moraEtAl2018} and \citet{dhuriEtAL2021}. \citet{moraEtAl2018} employed the PKLot in the tests, considering camera angles, climate conditions, and parking lot changes.
The authors also propose a method based on Bag of Features, discussed in Section \ref{subsec:featureExtractionMethods}. In \citet{dhuriEtAL2021} the PKLot and CNRPark-EXT datasets are used to test scenarios containing parking lot changes. The authors used a small custom dataset containing 1,000 cropped images of individual parking spaces as well.

The MobileNetV2 \citep{SandlerEtAl2018} was used to create a lightweight network for parking spaces classification in \citet{MerzougEtAl2019}. The authors employed the PKLot, CNRPark-EXT, and a private dataset during the tests. 
However, the authors' testing procedure is not clear. The results were reported as the number of detected vehicles without any ground truth for comparison. A shallower version of the ResNet50 \citep{heEtAl2016} residual network is used to classify the individual parking spaces in \citet{gregorEtAl2019}. The authors used the PKLot and a private dataset for the tests. The training/testing instances were split using stratified sampling. The original ResNet50 network is employed in \citet{baktirBolat2020}. They employed small subsets of the PKLot and CNRPark-EXT datasets for the tests.

In \citet{chenEtAl2020}, a lightweight version of the Yolov3 network that uses the MobileNetV2 extraction layer is employed for classification.
The authors consider that the images may come from a video stream source.
Thus the parking spot is considered occupied if the bounding box of a car detected by the network overlaps a parking space during a time window of $n$ images.
The CNRPark-EXT and a private dataset were used during the tests.

Custom models were proposed by \citet{diMauroEtAl2016, jensenEtAl2017, dimauro2017park, thomas2018smart, nurullayevLee2019, shahEtAl2020}.
A fine-tuned AlexNet network is used in \citet{diMauroEtAl2016} and \citet{dimauro2017park}, using images from the same parking lot and camera angle (from PKLot and a private dataset).
A pseudo-label model for semi-supervised learning is employed in \citet{diMauroEtAl2016} to compare with AlexNet, beating the former accuracy.
\citet{jensenEtAl2017} proposed a model with a fixed input size of 40x40 pixels in the PKLot dataset using the original test protocol proposed for the PKLot.
In \citet{nurullayevLee2019} is proposed a \ac{CNN}-based method using dilated convolutions that achieved promising results, which skips pixels in the convolution kernel. According to the authors, it increases the classifier's ability to learn the global context of the images.
\citet{thomas2018smart} also used a custom model, but they did not specify its structure or parameters.

The authors in \citet{shahEtAl2020} proposed a custom network called Fully-Multitask Convolutional Neural Network. The proposed approach depends on the Mask R-CNN for the extraction of masked regions. The authors claim that the proposed approach is capable of counting and automatically detecting the parking spaces. However, no details about these features nor results are given (thus, this work is not considered in Sections \ref{sec:parkingSpacesSegmentation} and \ref{sec:carsCounting}). For the individual parking spaces classification, the authors report quantitative results only for the training and validation phases in the PKLot dataset (results for the test phase are given in a plot according to the training epoch, making its analysis difficult).

\ac{DL} models for detection and segmentation were employed in \citet{nietoEtAl2019} and \citet{sairamEtAl2020} to aid vehicle detection. \citet{nietoEtAl2019} used Faster R-CNN \citep{renEtAl2015} and multiple cameras from their proposed \ac{PLds} dataset. 
They applied a homographic transformation and perspective correction to transform the plane of each camera to a common plane and correct the positions of the detected cars, respectively. Thereafter, the cars detected by the different cameras were fused to classify the parking spots.
The Faster R-CNN was also used in \citet{KhanEtAl2019}, where the authors focused on tests involving different camera angles and parking lot changes using the PKLot dataset.

More recently, \citet{sairamEtAl2020} proposed a method based on the Mask R-CNN \citep{kaimingEtAl2017} network. It was used to extract individual vehicles and to detect the proportion of the parking space the vehicles are occupying to differentiate between cars and two-wheel vehicles.
The Mask R-CNN is also used in \citet{agrawalEtAl2020}. The network is employed to detect the individual parking spaces (details in Section \ref{sec:parkingSpacesSegmentation}) and classify the spots between occupied and empty. The authors trained their model using the COCO \citep{lin2014microsoft}, and COWC datasets \citep{mundhenkEtAl2016Cowc}, and tested using the PKLot dataset. 

In \citet{mettupallyMenon2019}, the Mask R-CNN network is trained to classify the individual parking spaces. The authors included the images' timestamps and orientation to improve the classification results and used a custom subset of the PKLot dataset containing 6,100 segmented images for the tests.

\acp{GAN} \citep{isolaEtAl2017} are employed in \citet{liChuahBhattacharya2017} to detect occupied automatically and vacant parking spaces using a team of drones. 
The \ac{GAN} is trained using the labeled images of the PKLot dataset, wherein the division between and training and testing subsets was carried out according to the original PKLot protocol. In the test phase, the images rotated at a random angle between $[-10, +10]$ degrees were included to simulate the data obtained from the drones.

\subsection{Automatic Parking Space Detection}\label{sec:parkingSpacesSegmentation}

\begin{figure}[ht]
    \centering
    \includegraphics[width=0.42\textwidth]{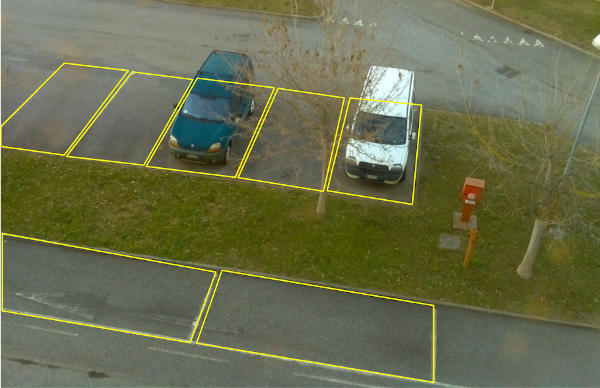}
    \caption{Parking spaces locations example(image from CNRPark-EXT).}
    \label{fig:parkingSpaceDetectionExample}
\end{figure}

The automatic parking space detection task focuses on automatically detecting the coordinates of each parking space, e.g., obtaining a bounding box, regardless of its status (occupied or empty).
This task is exemplified in Figure \ref{fig:parkingSpaceDetectionExample}, where the yellow polygons (that define each parking space) coordinates should be automatically defined.
It is a challenging task since parking spaces are similar to roads, i.e., how can a model discriminate between a parking space and a road segment?
The presence of cars may hinder correct detection, especially for methods that rely on the painted demarcations in the parking lots that delimits the parking spaces.

An automatic approach for detecting parking spaces using classical image processing methods is proposed in \citet{bohushEtAl2018}.
The approach uses perspective transformation in the entire input image. It makes the parking spaces rectangular and parallel to the axes (the approach may not be suitable for parking areas where the cars park in angled parking spaces).
Otsu's binarization and morphological operations are used for parking space detection, where painted lines must delimit the parking spaces. The authors did not make clear the results achieved by their proposed method.
\citet{ZhangEtAl2019} also proposed an approach for automatic parking space detection using perspective transformation and classical image processing approaches, such as the Canny and Gaussian edge detectors. The authors did not present quantitative results.

Further, the authors in \citet{vitekMelnicuk2018} employed a grid-based approach, where for each block of the grid, \ac{HOG} features are extracted. 
Thereafter, a classifier is used to classify each block as a car or street. Blocks classified as cars are merged into parking spaces according to their neighbor blocks. Adjacent blocks are classified as cars that are considered as belonging to the same parking spot. The method seems to detect cars and not necessarily parking spaces, e.g., a car may be just passing by the parking lot. The authors only report some images with qualitative results.

By assuming that the car park area is rectangular (forming a parking grid), the authors in \citet{nietoEtAl2019} can automatically define each parking spot. Given an aerial image of the parking lot (computed using a homography matrix and a regular image collected by a camera), the corners of the parking grid, and the number of rows and columns, the method can automatically define each parking space of the parking grid.

The \ac{GAN}-based approach by \citet{liChuahBhattacharya2017} generates parking spaces from the PKLot dataset using the manually-made masks to train the network.
Although parking spaces are segmented, only an evaluation of individual parking spaces is made. Evaluation of the automatic detection was not executed.
\citet{agrawalEtAl2020} proposed using the Mask R-CNN to identify the positions where the cars stay parked. The authors use this information to extract the parking space positions by assuming that areas where cars stay parked for long periods, can be considered parking spaces. Unfortunately, the authors do not describe the details of the parking space detection approach. Also, they do not give quantitative results.

The authors in \citet{padmasiriEtAl2020} used the ResNet \citep{heEtAl2016} and Faster-RCNN networks to detect the parking spaces in the PKLot dataset automatically. The authors reported results using the \ac{AP} metric. However, images from the same parking lot were used for both training and testing, which may lead to biased results, as discussed in Section \ref{subsec:parkinExtracionDiscussion}.

A two-step automatic parking space detection approach is proposed by \citep{patel2020car}. First, an object detector, Faster R-CNN (same work with this detector is found in \citet{kirtibhai2020faster}) or YOLOv4\citep{bochkovskiy2020yolov4}, is employed for car detection (trained in the CARPK dataset \citep{hsiehLinHsu2017}). Then, the bounding boxes detected are used for car tracking.
For car tracking, the idea is to find bounding boxes with a stationary car for some time/frames.
They evaluated the proposed two-step approach using only three busy days from CNRPark-EXT for each weather condition, introducing bias in the tests.
A parking space is marked if a car is parked for at least one hour and a half. 

\subsection{Car Detection and Counting}\label{sec:carsCounting}

\begin{figure}[ht]
    \centering
    \includegraphics[width=0.45\textwidth]{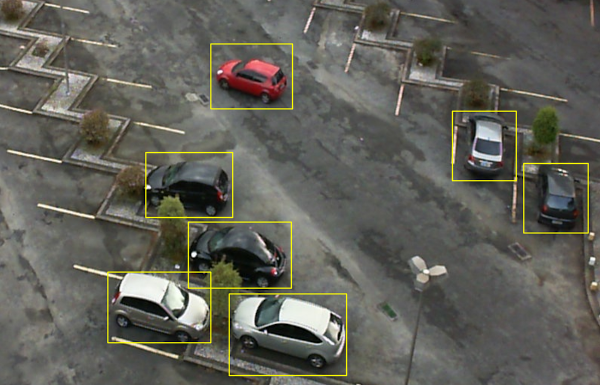}
    \caption{Car Detection example. Image containing 7 cars (image from PKLot).}
    \label{fig:carDetectionExample}
\end{figure}

In this Section, methods that aim to detect and count individual cars in the images are presented.
For car detection, bounding boxes or segmentation masks may be generated and evaluated for each car, as exemplified in Figure \ref{fig:carDetectionExample}.
In contrast, we have a regression problem for the car counting task. In this case, we are only interested in the discrete final number of cars present in the image (seven in the example in Figure \ref{fig:carDetectionExample}).
Like in the individual parking spaces classification problem, an image from the parking lot is the input. Different from it, no pre-processing to extract the individual parking spaces is applied.
Proposed works aim to achieve a high car detection precision and reduce the difference between the prediction and the actual number of cars in the images. 

Common datasets used to evaluate methods that deal with this task include the PUCPR+ (an extension of the PUCPR subset from the PKLot) \citep{hsiehLinHsu2017, liEtAl2019, dominikGabzdyl2020}, the original PUCPR subset of the PKLot \citep{laradjiEtAl2018}, the complete PKLot dataset \citep{vargheseSreelekha2019}, the CARPK dataset \citep{hsiehLinHsu2017, dominikGabzdyl2020}, and drone-acquired image datasets \citep{hsiehLinHsu2017, liEtAl2019}.

Most authors use \ac{DL}-based approaches for car detection. They also count the number of detected instances (e.g., bounding boxes) to obtain the final prediction for the number of cars in the images \citep{hsiehLinHsu2017, laradjiEtAl2018, liEtAl2019, amatoEtAl2018, amatoEtAl2019}.
The authors in \citet{hsiehLinHsu2017} aimed to detect and count aerial images from drones but used 5-fold cross-validation (which may overestimate the reported results). In contrast, \citet{liEtAl2019} generated the anchors for the training phase adaptively.

In \citet{laradjiEtAl2018} a new loss function with a convolutional model for car detection and counting is proposed. They also used annotations that roughly contain the objects of interest, which are, according to the authors, easier to label manually. The Yolov3 was used in \citet{amatoEtAl2019} to count the number of cars in the images. 
The authors used the PUCPR+ and CARPK datasets in the tests using the test protocols proposed by \citet{hsiehLinHsu2017}. Similarly, \citet{amatoEtAl2018} (also in \citet{ciampi2018counting}) used the Mask R-CNN for counting by density in a CNRPark-EXT counting version called Counting CNRPark-EXT.
Since the original dataset does not have car masks, they initially trained Mask R-CNN to correctly generate mask predictions of cars (using about 10\% of the training subset). Then, they retrained Mask R-CNN with the generated masks (and some cases that were manually corrected).

The authors in \citet{vargheseSreelekha2019} and \citet{SharmaPandey2021} proposed detection-only approaches. In the work of \citet{vargheseSreelekha2019} a background subtraction-based method is used for hypothesis generation of the possible areas where cars park. A shadow model is employed to reduce the amount of noise. Then a classifier is used to verify if the segmented areas contain cars. A custom \ac{CNN} is proposed in \citet{SharmaPandey2021}. The authors claim that the custom network is lightweight and can be processed in a regular CPU.

The approaches discussed in \citet{dominikGabzdyl2020,stahlEtAl2018,dobes2020} count the number of objects in the images globally, without employing a car detection step.
\citet{dominikGabzdyl2020} proposed a tree-like \ac{CNN}-based car counting approach. The first ten layers from a VGG16 network are used for feature extraction. The following convolutional layers are used to generate a density prediction, then used to count cars. Point-wise and dilated convolutions were employed in this part.

The authors in \citet{stahlEtAl2018} proposed an approach where the image is divided into several patches and fed to a modified R-FCN \citep{daiEtAl2016} network. The method includes an inclusion-exclusion layer to detect objects counted more than once since an object may appear in more than one patch. The approach only needs the number of objects in the training images for the training phase. It was evaluated under several object counting benchmarks, including the PUCPR+ dataset. A modified version of the Stacked Hourglass Network \citep{YangJia2016} is used in \citet{dobes2020}. The modified network can identify the best scale of the input images to count the cars.  The network input is a pyramid of gradually downsampled images and, for the training phase, the network needs labels in the form of point annotations.  The authors used the PUCPR+ altogether with other car counting benchmarks.

\section{Discussion}\label{sec:discussion}

In this section, the datasets reviewed and works that used them are summarized and discussed.
Figure \ref{subfig:mainTasksDistribution} shows that most works (\overallParkingClassificationOnly) focus on the individual parking spaces classification only.
As one can observe in Figure \ref{subfig:overallDatasetsUsage}, the PKLot dataset is the most popular dataset when considering the surveyed works, being used in 88\% (70\% + 18\%) of the works, while the CNRPark-EXT and the PLDs datasets are used in 29\% (11\% + 18\%) and 1\% of the works, respectively.
The difference between the usage ratios can be partially explained by the publication dates of these datasets, as PKLot, CNRPark-EXT, and \ac{PLds} datasets were released in 2015, 2017, and 2019, respectively.

{
    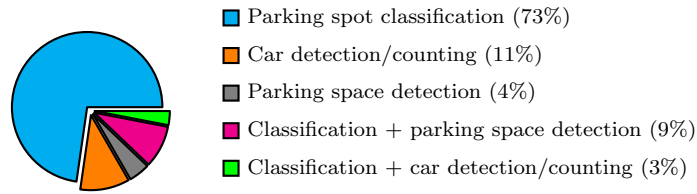
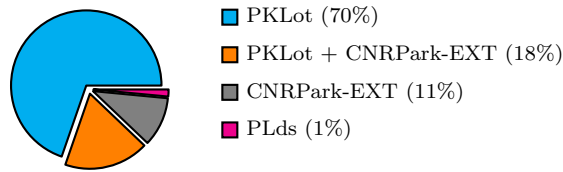
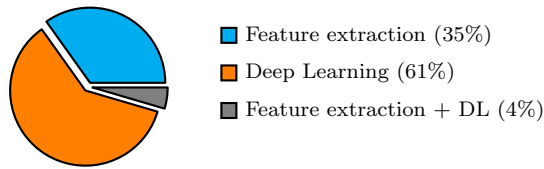
\begin{figure}[ht]
        \centering
        \subfloat[Main tasks considered in the surveyed works.]{
            \label{subfig:mainTasksDistribution}
            
    \tikzsetnextfilename{./tikz/typesTasks}%
    \begin{tikzpicture}
\togglefalse{showpct}
\footnotesize
 \pie [explode=0.06, text=legend, radius=1.0, sum=auto, color={cyan, orange, gray, magenta, green}]
    {   48/ Parking spot classification (73\%),
        7/ Car detection\text{/}counting (11\%),
        3/ Parking space detection (4\%),
        6/ Classification + parking space detection (9\%),
        2/ Classification + car detection\text{/}counting (3\%)}
\end{tikzpicture}%

        }\\
        \subfloat[Overall datasets usage.]{
            \label{subfig:overallDatasetsUsage}
            
    \tikzsetnextfilename{./tikz/overallDatasetsUsage}%
    \begin{tikzpicture}
\togglefalse{showpct}
\footnotesize
 \pie [explode=0.06, radius=1.0, text=legend, color={cyan, orange, gray, magenta}, sum=auto]
    {
    46/ PKLot (70\%),
    12/ PKLot + CNRPark-EXT (18\%),
    7/ CNRPark-EXT (11\%),
    1/ PLds (1\%)}
\end{tikzpicture}%

        }\\
        \subfloat[Feature extraction and Deep Learning usage.]{
            \label{subfig:overallfeatureExtractionVersusCNN}
            
    \tikzsetnextfilename{./tikz/overallfeatureExtractionVersusCNN}%
    \begin{tikzpicture}
\togglefalse{showpct}

\footnotesize
 \pie [explode=0.06, text=legend, radius=1.0, sum=auto, color={cyan, orange, gray}]
    {   23/ Feature extraction (35\%),
        40/ Deep Learning (61\%),
        3/ Feature extraction + DL (4\%)
    }
\end{tikzpicture}%

        }
        \caption{General findings considering all surveyed works.}
        \label{fig:plotsGeneralFindings}
    \end{figure}
}

In Figure \ref{subfig:overallDatasetsUsage}, it is also possible to notice that only 18\% of the surveyed works use more than one publicly available dataset for the tests. When comparing the proportion of approaches based on Feature Extraction and Deep Learning shown in Figure \ref{subfig:overallfeatureExtractionVersusCNN}, it is possible to check that \ac{DL}-based approaches are more prevalent in general.

An overview of our findings and recommendations regarding the surveyed datasets and research gaps for the individual parking space classification, parking space detection, and car detection/counting tasks are the following:
\begin{enumerate}
    \item The current publicly available datasets lack some features. Future datasets should be robust as defined in this work and may contain:
    \begin{itemize}
        \item Video sequences;
        \item Images in nighttime and snow conditions;
        \item Labels, other than the parking spaces, such as segmentation masks for the objects (e.g., cars, obstacles, pedestrians).
    \end{itemize}
    \item There is a lack of standard protocols for testing approaches:
        \begin{itemize}
            \item New protocols should include the data split procedure, evaluation metrics, and specific challenges (e.g., generalization problems, automatic parking space detection problems);
            \item The protocols should take into consideration realistic scenarios.
        \end{itemize}
    \item Many of the surveyed works are not reproducible:
        \begin{itemize}
            \item Authors should use publicly available datasets;
            \item Standard test protocols, as suggested in Item 2, should be used.
        \end{itemize}
    \item The individual parking spaces classification under camera or parking lot change scenarios is an open problem:
        \begin{itemize}
            \item Authors should consider a multiple datasets perspective, for instance, training in the PKLot dataset and testing the CNRPark-EXT.
        \end{itemize}
    \item The automatic parking space detection is an open problem:
    \begin{itemize}
            \item Quantitative metrics must be used to report the results;
            \item Standard test protocols should be created (as in Item 2);
            \item Authors should consider multiple datasets (as in Item 4).
        \end{itemize}
\end{enumerate}
These findings are discussed more deeply in Sections \ref{subsec:datasetsSummary}, \ref{subsec:discussIndividualParkingLot}, \ref{subsec:parkinExtracionDiscussion}, and \ref{subsec:carCountDiscussion}.
In Section \ref{subsec:datasetsSummary}, we summarize and discuss the surveyed datasets.
In Sections \ref{subsec:discussIndividualParkingLot}, \ref{subsec:parkinExtracionDiscussion}, and \ref{subsec:carCountDiscussion}, we present, compare, and discuss the authors' results in the individual parking spaces classification, parking space detection, and car detection/counting tasks, respectively.
These discussions are used to provide a thoughtful understanding of the vision-based parking lot management problems that already have a factual basis for solutions and open problems that require more research and attention from the scientific community.

\subsection{Parking Lot Datasets} \label{subsec:datasetsSummary}

Table \ref{table:comparacaoDatasets} shows the main features of the datasets discussed in this paper. Images are classified according to the climate condition, regardless if they were acquired under day or nighttime. For the overcast climate condition during nighttime, we considered the images right after rains when the floor is visibly wet and may contain water puddles.

\begin{table*}[ht]
\caption{Main features of the datasets}
\centering
\tiny
\setlength{\tabcolsep}{3px}
\begin{threeparttable}[c]
    \begin{tabular}{L{35px}rrrrR{20px}rrr}
        \multicolumn{9}{l}{\textbf{PKLot} -- $1280 \times 720$ pixels, 12,417 daytime images} \\\hline
        \multirow{2}{35px}{Camera/ Park. Lot} & \multicolumn{4}{c}{\# Of Days (\# Of Images) } & \multirow{2}{20px}{spaces/ image} & \multicolumn{3}{c}{Labeled Parking Spaces} \\
        \cline{2-5}\cline{7-9}
        & Clear Sky & Overcast   & Rain & Snow  &   & Occupied & Empty & Total \\\hline
        
        UFPR04 & 20 (2,098) & 15 (1,408) & 14 (285) & - & 28 & 46,125 (43.6\%) & 59,718 (56.4\%) & 105,843 \\
        UFPR05& 25 (2,500) & 19 (1,426) & 8 (226) & - & 45 & 97,426 (58.8\%) & 68,359 (41.2\%) & 165,785 \\
        PUCPR  & 24 (2,315) & 11 (1,328) & 8 (831) & - & 100 & 194,229 (45.8\%) & 229,994 (54.2\%) & 424,223 \\
        Total & 69 (6,913) & 45 (4,162) & 30 (1,342) & - & - & 337,780 (48.6\%) & 358,071 (51.4\%) & 695,851\\\hline
        \multicolumn{9}{l}{\textbf{CNRPark-EXT} -- $1000 \times 750$ pixels, 4,278 daytime images} \\\hline
        Camera A  & 2 ( - ) & 0 ( - ) & 0 ( - )&   &  & 3,621 (58.7\%) & 2,550 (41.3\%) & 6,171\\
        Camera B  & 2 ( - )& 0 ( - ) & 0 ( - )&   &  & 4,781 (74.6\%)& 1,632 (25.4\%) & 6,413\\
        Camera 1  & 10 (198)& 7 (137)& 6 (124)& - & 35 & 9,308 (59.2\%)& 6,407 (40.8\%)& 15,715\\
        Camera 2  & 10 (201)& 7 (139)& 6 (124)& - & 10 & 2,641 (64.5\%)& 1,454 (35.5\%)& 4,095\\
        Camera 3  & 10 (198)& 7 (138)& 6 (124)& - & 22 & 5,370 (56.7\%)& 4,101 (43.3\%)& 9,471\\
        Camera 4  & 10 (197)& 7 (137)& 6 (123)& - & 38 & 9,357 (56.4\%)& 7,219 (43.6\%)& 16,576\\
        Camera 5  & 10 (204)& 7 (143)& 6 (127)& - & 47 & 11,256 (54.0\%)& 9,582 (46.0\%)& 20,838\\
        Camera 6  & 10 (211)& 7 (154)& 6 (130)& - & 45 & 10,646 (52.9\%)& 9,462 (47.1\%)& 20,108\\
        Camera 7  & 10 (211)& 7 (154)& 6 (130)& - & 47 & 10,519 (49.8\%)& 10,595 (50.2\%)& 21,114\\
        Camera 8  & 10 (206)& 7 (151)& 6 (127)& - & 55 & 12,847 (53.3\%)& 11,237 (46.7\%)& 24,084\\
        Camera 9  & 10 (205)& 7 (154)& 6 (131)& - & 29 & 7,363 (56.8\%)& 5,601 (43.2\%) & 12,964\\
        Total\tnote{1} & 12 (1,831) & 7 (1,307)& 6 (1,140) & - & - & 87,709 (55.7\%)& 69,840 (44.3\%)& 157,549\\\hline
        \multicolumn{9}{l}{\textbf{PLds} -- $1280 \times 960$ pixels, 5,215 daytime images (62.5\%), 3,125 (37.5\%) nighttime images} \\\hline
        isshk & 45 (2,280) & 22 (635) &21 (428) & 8 (321) & - & 32,254 (100\%) & - & 32,254\\
        vxusd/vmlix\tnote{2} & 46 (2,077) &  37 (1,018) & 27 (485) & 2 (96) & - & 26,719 (100\%)& - & 26,719\\
        qridr & 18 (435) & 15 (513) & 6 (52) & - & - & - & - & - \\
        Total\tnote{1} & 69 (4,792) & 47(2,166) & 34 (965) & 10 (417) & - & 58,973 (100\%) & - & 58,973\\\hline
    \end{tabular}
    \begin{tablenotes}
        \item[1] Different cameras can collect images in the same days, thus the total number of days may not be equal to the sum of the days for each camera; 
        \item[2] We detected that the vxusd and vmlix refer to the same camera angle.
    \end{tablenotes}
\end{threeparttable}
\label{table:comparacaoDatasets}
\end{table*}

Table \ref{table:comparacaoDatasets} indicates that the PKLot is the dataset with the most parking lot images (12,417) and individual parking spaces samples available (695,851).
Another interesting feature of the PKLot, unavailable in other surveyed benchmarks, is having more than one parking lot (i.e., PUCPR and UFPR).
Regarding the CNRPark-EXT, one of the most appealing features is its vast number of camera angles available, which can be used in camera angle change tests.
In Table \ref{table:comparacaoDatasets}, it is also possible to check that both the PKLot and CNRPark-EXT datasets are reasonably well balanced between the occupied and empty classes.

When considering the PLds dataset, we observe that it is the only surveyed dataset that contains images collected in snow and nighttime conditions. Another attractive feature of the PLds dataset is the presence of images synchronized between different cameras. However, only 100 synchronized images are available. The main drawback of the PLds dataset is the absence of annotated empty parking spaces.

Although all the datasets combined give 25,035 parking lot images and 912,373 individual parking spaces images, only 417 (1.7\%) images represent snow conditions, and 3,125 (12\%) images were collected under nighttime.
Both snow and nighttime condition images are from the PLDs dataset and were collected in a single parking lot.
None of the surveyed datasets contained video sequences. The only labels available were the climate condition, the positions of the parking spaces and their statuses (for the PKLot and CNRPark-Ext), and the parked cars bounding boxes (for the \ac{PLds}).

These findings indicate that new datasets are needed, mainly containing conditions such as nighttime and snow climate. These datasets may also contain video sequences and labels regarding objects other than the parking spaces, such as segmentation masks for the objects in the scene (e.g., vehicles, pedestrians, obstacles). These datasets could help create more general methods for the tasks described in this paper and other tasks, such as object tracking.

\begin{figure}[htpb]
    \centering
     \subfloat[17:12]{
        \centering
        \includegraphics[width=0.08\textwidth]{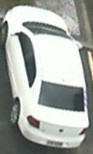}
    }\hspace{0.1cm}
     \subfloat[17:32]{
        \centering
        \includegraphics[width=0.08\textwidth]{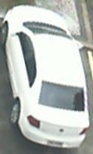}
    }\hspace{0.1cm}
     \subfloat[17:52]{
        \centering
        \includegraphics[width=0.08\textwidth]{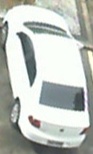}
    }\hspace{0.1cm}
     \subfloat[18:12]{
        \centering
        \includegraphics[width=0.08\textwidth]{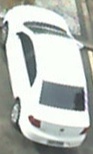}
    }\hspace{0.1cm}
     \subfloat[18:32]{
        \centering
        \includegraphics[width=0.08\textwidth]{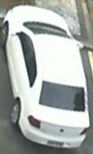}
    }
    \caption{Car parked for several minutes in a parking space (PKLot).}
    \label{fig:sameCar}
\end{figure}

When considering the test protocols, the authors of all discussed datasets suggested that the same day's images may belong to just one set - training or test set.
This rule avoids pictures related to the same car parked in the same space for hours to appear in the training and the testing sets simultaneously, creating trivial and unrealistic problems -- See an example in Figure \ref{fig:sameCar}.
\citet{almeidaEtAl2015} suggest that the PKLot dataset must be split using 50\% of the days for training and the other 50\% for tests.
\citet{amatoEtAl2017} gives their train and test images for the CNRPark-EXT dataset, containing 74.8\% and 25.2\% of the images, respectively.
The authors of the \ac{PLds} dataset \citep{nietoEtAl2019} also made available the images already partitioned between the train and test sets, containing 76\% and 24\% of the images, respectively.

In \citet{almeidaEtAl2015} and \citet{amatoEtAl2017}, the test protocols include camera angle and parking lot changes that may represent more challenging and realistic situations.
The general ideas of the tests suggested by the authors of the datasets are next described. These scenarios (single parking lot, camera angle change, and parking lot change) are considered in Section \ref{subsec:discussIndividualParkingLot}. There, the surveyed techniques regarding the individual parking spot classification problem are compared.

\begin{itemize}
    \item Single parking lot: images for both training and test sets come from the same parking lot and are collected using the same camera angle;
    
    \item Camera angle change: images must be from the same parking lot, but the train set must contain images collected using a different camera angle from the test set's images;
    
    \item Parking lot change: the test set must contain images from a parking lot different from the train set.
\end{itemize}

\subsection{Individual Parking Spot Classification}\label{subsec:discussIndividualParkingLot}

Table \ref{table:comparacaoMetodosFeature} shows a summary of the individual parking spot classification methods based on feature extraction approaches. As for \ac{DL}-based methods, a summary can be seen in Table \ref{table:comparacaoMetodosDeep}. Works that do not clearly describe the test details are marked as not reproducible in these tables. This marking indicates that not enough details are given for other researchers to reproduce the experiments. Some of the main reasons encountered in the works that led us not to consider the results reproducible include the use of images in private datasets, use of not well-described subsets of the public datasets, and incomplete evaluation protocols, e.g., the protocol used to split the dataset between train and test is not given. Even though all surveyed methods use at least one public dataset, surprisingly, 56\% of them were deemed as not reproducible.

Different authors may use distinct metrics to assess their methods, such as accuracy, \ac{AUC-ROC}, and \ac{AUC-PR}. This pattern imposed a challenge to summarize the methods, and we chose to show all results in terms of accuracy rates. The accuracy may not be the most suitable metric to assess the parking spaces classification task, e.g., due to class imbalances in the test sets. Nevertheless, we use it to show the results since most authors reported their results using this metric. 
On the other hand, for specific scenarios in which accuracy was not used, we report the results following the metric adopted in the original studies depicted in Tables \ref{table:comparacaoMetodosFeature} and \ref{table:comparacaoMetodosDeep}. 
Authors that did not present quantitative results in their works are not discussed in this section.

\begin{table}[htpb]
\caption{Feature extraction based individual parking spaces classification approaches overview.}
\centering
\tiny
\setlength{\tabcolsep}{2px}
\begin{threeparttable}[c]
    \begin{tabular}[c]{L{65px}L{70px}L{22px}R{38px}R{38px}R{38px}L{53px}}
        \hline
        &  &  & \multicolumn{3}{c}{Accuracy (\%)} &\\\cline{4-6}
        Authors, year & Classifier and Features &Repro-ducible& \multicolumn{1}{C{38px}}{Single Parking Lot} & \multicolumn{1}{C{38px}}{Angle Change} & \multicolumn{1}{C{38px}}{Parking Lot Change} & Datasets Used\\\hline
        
        \citet{almeidaEtAl2013} & Ensembles of \acs{SVM} and \acs{LBP}/\acs{LPQ} & yes & 99.8\%\tnote{1} & - & - & PKLot (UFPR04)\\
        
        \citet{almeidaEtAl2015} & Ensembles of \acs{SVM} and \acs{LBP}/\acs{LPQ} & yes & 99.3 - 99.6\% & 85.8 - 88.3\% & 84.2 - 89.8\% & PKLot\\
        
        \citet{baroffioEtAl2015} & \acs{SVM} and \acs{HSV} Histogram & yes & 87.0 - 96.0\% & - & - & PKLot \\
        
        \citet{AhrnbomAstromNilsson2016} & \ac{SVM}/\acs{LR} and Feature Channels & yes & 0.999 - 1.0\tnote{3} & 0.978 - 0.996\tnote{3} & 0.940 - 0.988\tnote{3}& PKLot\\
        
        \citet{suwignyoSetyawanYohanes2018} & \acs{k-NN}/\acs{SVM} and \acs{QLRBP} & yes & 99.2 - 99.4\%\tnote{1}  & - & - & PKLot \\
        
        \citet{dizon2017development} & Background subtraction with \acs{AMF} and linear \acs{SVM}/\acs{LBP} with \acs{HOG} & no & 1.8\%/21.0\%\tnote{4} & - & - & PKLot (UFPR04)\tnote{2}\\
        
        \citet{almeidaEtAl2018} & Dynse Framework and \ac{LBP}& yes & 92.2\% & - & - & PKLot \\
        
        \citet{bohushEtAl2018} & \acs{SVM} and \acs{HOG} & no & 99.7\%\tnote{1} & - & - & PKLot (Subset) \\
        
        \citet{vitekMelnicuk2018} & \acs{SVM} and \acs{HOG} & no & 90.7 - 93.2\%\tnote{1} & - & 83 - 96\% & PKLot\tnote{2} \\
        
        \citet{hammoudiEtAl2018} & \ac{k-NN} and \ac{LBP}-based features & no & 98\% - 98.6\%\tnote{1} & - & - & PKLot(Subset)\\
        
        \citet{hammoudiEtAl2018b} & \ac{k-NN} and \ac{LBP}-based features & no & 88.65\tnote{1} & - & - & PKLot(Subset)\\
        
        \citet{moraEtAl2018} & \ac{SVM} and SIFT (Bag of Features) & yes & 76.8 - 92.0\%\tnote{1} & 94.1 - 99.7\% & 90.4 - 97.3\% & PKLot\\
        
        \citet{amatoEtAl2018} & Background modeling and Canny & yes & 99.4\% & - & - & CNRPark-EXT \\
        
        \citet{dornaikaEtAl2019} & \acs{SVM}/\acs{k-NN} and textural features & yes & 88.7 - 99.6\%\tnote{1} & - & 63.0 - 87.8\% & PKLot, CNRPark-EXT\\
        
        \citet{raj2019vacant} & Canny and color features with Random Forest & yes & 98.31\%\tnote{1} & - & - & PKLot\\
        
        \citet{hammoudiEtAl2019} & \ac{k-NN}/\ac{SVM} and \ac{LBP}-based features & no & 85.6\%\tnote{1} & - & 71.0 - 80.5\% & PKLot, CNRPark-EXT (Subsets)\\
        
        \citet{thike2019parking} & Threshold and \ac{ULBP} & no & 79\%\tnote{1} & - & - & PKLot\\
        
        \citet{vargheseSreelekha2019} & \acs{SVM} and \acs{SURF} & yes & 91.1 - 99.7\% & - & 81.7\% & PKLot, CNRPark-EXT\\
        
        \citet{farag2020deep} & Threshold and color average & no & 79.0 - 90.0\%\tnote{1} & - & - & PKLot, CNRPark-EXT \\
        
        \citet{irfanEtAl2020} & Euclidean Similarity and \acs{GLCM} & no & 95.0\% & - & - & PKLot(PUCPR)\tnote{2}\\
        
        \citet{hammoudiEtAl2020} & \ac{k-NN} and \ac{LBP}-based features & no & 87.0\%\tnote{1} & - & - & PKLot(UFPR04)\\
        
        \citet{magoKumar2020} & Neural Networks, \ac{k-NN}, \ac{SVM}, Naïve Bayes, textures and color features & no & 99.7\%\tnote{1} & - & - & PKLot \\
        
        \citet{almeidaElAl2020} & Approaches for concept drift and \ac{LBP} features& yes & 86.7 - 90.4\% & - & - & PKLot \\
        
        \hline
    \end{tabular}
    \begin{tablenotes}
        \item[1] The test procedure may have led to biased results, or it is not clear. 
        \item[2] Authors included private datasets in the tests.
        \item[3] Results in \ac{AUC-ROC}.
        \item[4] Results in \ac{FNR}/\ac{FPR}.
    \end{tablenotes}
\end{threeparttable}
\label{table:comparacaoMetodosFeature}
\end{table}

The results in Tables \ref{table:comparacaoMetodosFeature} and \ref{table:comparacaoMetodosDeep} are reported according to the authors' tests, including scenarios where the images from the training and test subsets come from the same parking lot and camera angle, scenarios containing camera angle changes, and scenarios when the train and test parking lots are different -- see Section \ref{subsec:datasetsSummary} for more details. For most works, we reported a range in the results in the format $x - y$, where $x$ is the worst result achieved in a given scenario, and $y$ is the best result. For instance, in the \textit{Single Parking lot} test, if a method achieved $88\%$, $97\%$, and $93\%$ of accuracy when tested in the UFPR04, UFPR05, and PUCPR PKLot's subsets, respectively, the reported result is $88 - 97\%$. It is worth mentioning that the results reported in Tables \ref{table:comparacaoMetodosFeature} and \ref{table:comparacaoMetodosDeep}  are not directly comparable since the authors may vary the datasets and experimental protocols.

\begin{table}[htpb]
\centering
\tiny
\setlength{\tabcolsep}{2px}
\caption{Deep learning based individual parking spaces classification approaches overview.}
\begin{threeparttable}[c]
    \begin{tabular}{L{65px}L{60px}L{22px}R{38px}R{38px}R{38px}L{53px}}
        \hline
        &  &  & \multicolumn{3}{c}{Accuracy (\%)}                       &\\\cline{4-6}
        Authors, year & Network &Repro-ducible& \multicolumn{1}{C{38px}}{Single Parking Lot} & \multicolumn{1}{C{38px}}{Angle Change} & \multicolumn{1}{C{38px}}{Parking Lot Change} & Datasets Used\\\hline
        
        \citet{amatoEtAl2016} & mAlexNet & yes & 89.8 - 99.6\% & 86.3 - 90.7\% & 82.9 - 90.4\% & PKLot, CNRPark-EXT (Cam. A and B)\\
        
        \citet{valipourEtAl2016} & VGGNet-F & yes &  99.6 - 100\% & 85.5 - 98.2\% & 92.2 - 99.2\% & PKLot\\
        
        \citet{diMauroEtAl2016} & AlexNet & no & 99.2 - 99.8\%\tnote{1} & - & - & PKLot\tnote{2}\\
        
        \citet{dimauro2017park} & AlexNet & no & 99.9 - 100\%\tnote{1} & - & - & PKLot\tnote{2}\\
        
        \citet{amatoEtAl2017} & mAlexNet & yes & 90.1 - 98.1\% & 93.3 - 93.7\% & 92.7 - 98.3\% & PKLot, CNRPark-EXT\\
        
        \citet{jensenEtAl2017} & Custom CNN & yes & 99.7 - 99.9\% & 95.5 - 96.0\% & 96.7 - 98.7\% & PKLot\\
        
        \citet{nyambalKlein2017} & AlexNet and LeNet & no & 98.0 - 99.0\% & - & - & PKLot\tnote{2}\\
        
        \citet{liChuahBhattacharya2017} & \acs{GAN} & yes & 94.7 - 97.5\%\tnote{3} & 56 - 94\%\tnote{3} & - & PKLot \\
        
        \citet{AcharyaYanKhoshelham2018} & VGGNet-F(feat. extr.)/\acs{SVM} & no & 99.7\%\tnote{1} & - & - & PKLot\tnote{2}\\
        
        \citet{buraEtAl2018} & Custom AlexNet & no & 99.5\%\tnote{1} & - & - & PKLot, CNRPark-EXT (Subsets)\tnote{2}\\
        
        \citet{thomas2018smart} & Custom CNN & no & 100\%\tnote{1} & - & - & PKLot\\
        
        \citet{amatoEtAl2018} & mAlexNet & yes & 99.6\% & - & - & CNRPark-EXT\\
        
        \citet{nurullayevLee2019} & CarNet & yes & 95.6 - 98.8\% & 95.2 - 97.6\% & 94.4 - 98.4\% & PKLot, CNRPark-EXT\\
        
        \citet{polprasertEtAl2019} & mAlexNet & no & 88\%\tnote{1} & - & - & PKLot\tnote{2}\\
        
        \citet{gregorEtAl2019} & Custom ResNet & no & 99.9\%\tnote{1} & - & - & PKLot\tnote{2}\\
        
        \citet{nietoEtAl2019} & Faster R-CNN & yes &  0.919\tnote{4} & - & - & PLds\\
        
       \citet{ZhangEtAl2019} & Custom VGG16 and Custom CNN & no & 99.1 - 99.9\%\tnote{1} & - & - & PKLot\\
        
        \citet{dingEtAl2019} & Yolov3 & no & 93.3\%\tnote{3} & - & - & PKLot \\
        
        \citet{KhanEtAl2019} & Faster R-CNN & yes & - & 82.5 - 96.4\% & 90.5 - 99.9\% & PKLot\\
        
        \citet{mettupallyMenon2019} & Mask R-CNN & no & 91.9\% \tnote{1} & - & - & PKLot\\
        
        \citet{agrawalEtAl2020} & Mask R-CNN & yes & - & 95\% & 95\% & PKLot \\
        
        \citet{baktirBolat2020} & ResNet50 & no & 99.5 - 99.8\%\tnote{1} & - & 97.4 - 99.3\% & PKLot, CNRPark-EXT (Subsets)\\
        
        \citet{sairamEtAl2020} & Mask R-CNN & no & 92\%\tnote{1} & - & - & PKLot\\
        
        \citet{chenEtAl2020} & Yolov3 & no & 99\%\tnote{1} & - & - & CNRPark-EXT\tnote{2}\\
        
        \citet{rahman2020convolutional} & CmAlexNet & no & 99.12\%\tnote{1} & - & - & CNRPark\tnote{1}\\
        
        \citet{AliMohamed2021} & Custom AlexNet & no & 92.9 - 98.9\%\tnote{1} & 94.9 - 96.9\% & 96.0 - 99.0\% & PKLot\\
        
        \citet{dhuriEtAL2021} & VGG16 & yes & 95.7 - 100\% & 97.1\% & 87.2\% & PKLot, CNRPark \\
        
        \citet{nguyen2021adaptive} & mAlexNet, AlexNet and MobileNet & no & 95.5 - 97.3\%\tnote{1} & - & - & CNRPark\tnote{2} \\
        
        \citet{KolharAlameen2021} & mAlexNet & no & 90.1 - 99.5\%\tnote{1} & - & - & PKLot, CNRPark-EXT\\ \hline
        
    \end{tabular}
    \begin{tablenotes}
        \item[1] The test procedure may have led to biased results or, it is not clear. 
        \item[2] Authors included private datasets in the tests.
        \item[3] Results in Precision.
        \item[4] Results in \ac{AUC-PR}.
    \end{tablenotes}
\end{threeparttable}
\label{table:comparacaoMetodosDeep}
\end{table}

Notice in Tables \ref{table:comparacaoMetodosFeature} and \ref{table:comparacaoMetodosDeep} that many authors, such as \citet{almeidaEtAl2013,diMauroEtAl2016,suwignyoSetyawanYohanes2018,dornaikaEtAl2019,vitekMelnicuk2018,AcharyaYanKhoshelham2018,polprasertEtAl2019,farag2020deep}, may have used test protocols that led to biased results.
Since a car may remain parked in the same spot in a parking lot for long periods, it is expected that the same car to appear in several consecutive images.
Any test approach that, for instance, randomly splits the data for training and testing may include images of the same car in both sets (training and testing) acquired at different times, leading to a trivial and unrealistic problem.
Figure \ref{fig:sameCar} shows an example where the same car remained parked in the same spot for several minutes.

All consecutive images acquired from a parked car should be exclusively selected as part of the training or testing sets in a more realistic scenario. The test protocols originally suggested in the datasets discussed in Section \ref{subsec:datasetsSummary} consider this problem. Protocols that ignore the fact that a car can stay parked in the same parking spot for a long time into consideration may have over-optimistic results. We reported and highlighted results from these approaches in Tables \ref{table:comparacaoMetodosFeature} and \ref{table:comparacaoMetodosDeep}.

In Table \ref{table:averagedResultsIndividualPkSpot}, the results reported by the different authors are averaged.
Only works deemed reproducible and not marked as possibly containing biased results were considered for this average.
As one can observe, on average, the accuracy reported for the single parking lot tests is of \globalAverageAccSinglePK.
Nevertheless, when considering the camera and parking lot change scenarios, the results are not as good, on average, reaching accuracies of \globalAverageAccCamChange\ and \globalAverageAccPkChange, respectively.
In \citet{almeidaEtAl2015}, where the PKLot is proposed, the authors recognized that the classifiers should be less dependent on the training set and proposed test protocols for this type of scenario. Nevertheless, as one can observe in Figure \ref{subfig:individualClassificationTestChanges}, most authors focus only on classifying images without considering any change, which is a more manageable problem.

\begin{table}[htb]
    \centering
    \footnotesize
    \caption{Averaged accuracy rates for the feature extraction and deep learning-based approaches. The numbers in parenthesis indicate the number of works considered in the averages.}
    \begin{tabular}{lrrr}
        Approach      & Single Park. Lot & Angle Change & Park. Lot Change \\\hline
        Feature extraction-based & 94.4\% (6)         & 87.0\% (1)   & 84.4\% (2) \\
        Deep learning-based & 97.6\% (7)         & 93.4\% (8)   & 93.7\% (8) \\\hline
        Average       & \globalAverageAccSinglePK (13)& \globalAverageAccCamChange (9)   & \globalAverageAccPkChange (10)
    \end{tabular}
    
    \label{table:averagedResultsIndividualPkSpot}
\end{table}

{
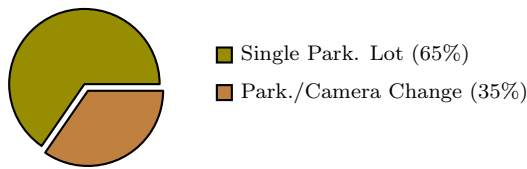
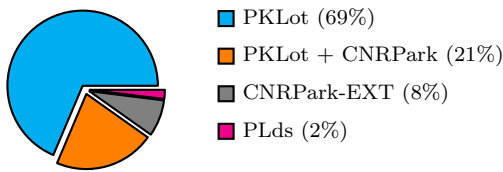
\begin{figure*}[htb]
        \centering
        \subfloat[Types of tests.]{
            \label{subfig:individualClassificationTestChanges}
            \centering
            
    \tikzsetnextfilename{./tikz/typesTestsIndividualParkingSpaces}%
    \begin{tikzpicture}
\togglefalse{showpct}
\footnotesize
 \pie [explode=0.05, text=legend, radius=1.0, color={olive, brown}, sum=auto]
    {   34/ Single Park. Lot (65\%),
        18/ Park.\text{/}Camera Change (35\%)}
\end{tikzpicture}%

        }
        \\
        \subfloat[Datasets usage.]{
            \label{subfig:datasetsUsageIndividualPK}
            \centering
            
    \tikzsetnextfilename{./tikz/datasetsUsageIndividualParkingClass}%
    \begin{tikzpicture}
\togglefalse{showpct}
\footnotesize
 \pie [explode=0.06, radius=1.0, text=legend, color={cyan, orange, gray, magenta}, sum=auto]
    {
    35/ PKLot (69\%),
    11/ PKLot + CNRPark (21\%),
    4/ CNRPark-EXT (8\%), 
    1/ PLds (2\%)}
\end{tikzpicture}%

        }
        
        \caption{Types of tests and datasets usage for the individual parking spaces classification problem.}
        \label{fig:typesAndDatasetsTestsIndividualPK}
    \end{figure*}
}

The datasets usage analysis presented in Figure \ref{subfig:datasetsUsageIndividualPK} also shows that only 21\% of the works employ more than one dataset in the testing procedure. Still considering Figure  \ref{subfig:datasetsUsageIndividualPK}, it is possible to verify that the PKLot is the most popular dataset, used in 90\% of the works (69\% + 21\%). The CNRPark-EXT dataset is used in 29\% of the works (8\% + 21\%). The PLds is used in 2\% of the works.
PKLot and CNRPark-EXT, and \ac{PLds} datasets were released in 2015, 2017, and 2019, respectively. 
As mentioned earlier, the high skew in the datasets usage may be (in part) due to the year of publication of each dataset.

When comparing the average results for approaches based on feature extraction and deep learning in Table \ref{table:averagedResultsIndividualPkSpot}, the results favor the \ac{DL}-based approaches.
However, the comparison made in Table \ref{table:averagedResultsIndividualPkSpot} should be taken with care since only a handful of works were used to compute the average. For example, only one feature extraction-based work was considered for the camera angle change scenario. When considering the number of works published, the results in Figure \ref{subfig:proportionFeatureDeep} show that the publications are almost evenly split between feature extraction and \ac{DL}-based approaches.

{
    \begin{figure*}[htb]
        \centering
        
    \tikzsetnextfilename{./tikz/legendaSubfigClassificadores}%
    \begin{tikzpicture}
\begin{axis}[%
    hide axis,
    xmin=0,
    xmax=1,
    ymin=0,
    ymax=1,
    legend style={legend cell align=left,align=left,draw=black,legend columns=2, column sep=2.5pt, draw=none}
]

\addlegendimage{area legend,color=black,solid, fill=cyan}
\addlegendentry{Feature extraction-based};

\addlegendimage{area legend,color=black,solid, fill=orange}
\addlegendentry{Deep learning-based};

\end{axis}
\end{tikzpicture}%

        \\
        \subfloat[Feature Extraction vs. Deep Learning.]{
            \label{subfig:proportionFeatureDeep}
            \centering
            
    \tikzsetnextfilename{./tikz/featureVSDeepIndividualParking}%
    \begin{tikzpicture}
\toggletrue{showpct}
\small
 \pie [explode=0.05, radius=1.24,color={cyan, orange}]
    {44/,
      56/}
\end{tikzpicture}
%

        }\hspace{0.2cm}
        \subfloat[Number of approaches using each feature extractor.]{
            \label{subfig:typesOfFeaturesExtracted}
            \centering
            
    \tikzsetnextfilename{./tikz/featureExtractorsIndividualParking}%
    \begin{tikzpicture}
    \footnotesize,
	\begin{axis}[
		xbar,
		width = 3.6cm,
        height = 4.2cm,
        bar width = 2mm,
		y axis line style = { opacity = 0 },
		tickwidth = 0pt,
		ytick = data,
		symbolic y coords = {LBP+Variants,Color Features,HOG,LPQ,Canny,Custom Features,GLCM,SIFT,SURF},
		xticklabel={\pgfmathparse{\tick}\pgfmathprintnumber{\pgfmathresult}},
		nodes near coords,
	]
	\addplot[black,fill=cyan] coordinates {
		(12,LBP+Variants)
        (5,Color Features)
        (3,HOG)
        (2,LPQ)
        (2,Canny)
        (1,Custom Features)
        (1,GLCM)
        (1,SIFT)
        (1,SURF)
	};
	\end{axis}
\end{tikzpicture}%

        }\hspace{0.2cm}
        \subfloat[Number of approaches using each classifier.]{
            \label{subfig:numApproachesClassifiers}
            \centering
            
    \tikzsetnextfilename{./tikz/classifiersUsedIndividualParking}%
    \begin{tikzpicture}
    \footnotesize,
	\begin{axis}[
		xbar,
		width = 3.9cm,
        height = 4.0cm,
        bar width = 2mm,
		y axis line style = { opacity = 0 },
		tickwidth         = 0pt,
		ytick = data,
		symbolic y coords = {SVM,K-NN,Background subtr.,Threshold,Drift Approaches,Logistic Regr.,Random Forest, Neural Network, Naive Bayes},
		xticklabel={\pgfmathparse{\tick}\pgfmathprintnumber{\pgfmathresult}},
		nodes near coords,
	]
	\addplot[black,fill=cyan] coordinates {
		(14,SVM)
        (8,K-NN)
        (2,Background subtr.)
        (2,Threshold)
        (2,Drift Approaches)
        (1,Logistic Regr.)
        (1,Random Forest)
        (1,Neural Network)
        (1,Naive Bayes)
	};
	\end{axis}
\end{tikzpicture}%

        }\hspace{0.2cm}
        \subfloat[Number of approaches using each network.]{
            \label{subfig:numApproachesNetwork}
            \centering
            
    \tikzsetnextfilename{./tikz/networksUsedIndividualParking}%
    \begin{tikzpicture}
    \footnotesize,
	\begin{axis}[
		xbar,
		width = 3.0cm,
        height = 4.1cm,
        bar width = 2mm,
		y axis line style = { opacity = 0 },
		tickwidth         = 0pt,
		ytick = data,
		symbolic y coords = {AlexNet+Variants,Custom Net.,VGG Net.,Mask R-CNN, Faster R-CNN,Yolov3,ResNet,LeNet,GAN,MobileNet},
		xticklabel={\pgfmathparse{\tick}\pgfmathprintnumber{\pgfmathresult}},
		nodes near coords,
	]
	\addplot[black,fill=orange] coordinates {
        (12,AlexNet+Variants)
        (4,Custom Net.)
        (4,VGG Net.)
        (3,Mask R-CNN)
        (2,Faster R-CNN)
        (2,Yolov3)
        (2,ResNet)
        (2,LeNet)
        (1,GAN)
        (1,MobileNet)
	};
	\end{axis}
\end{tikzpicture}%

        }
        
        \caption{Individual Parking Spaces Classification Problem -- Proportion and number of works taking into consideration different approaches. Works that present multiple approaches or use more than one feature/classifier may be computed more than once.}
        \label{fig:C45PklotTest}
    \end{figure*}
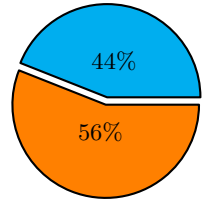
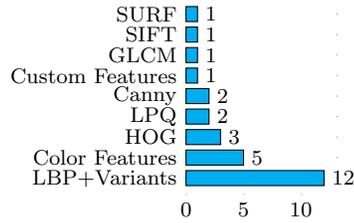
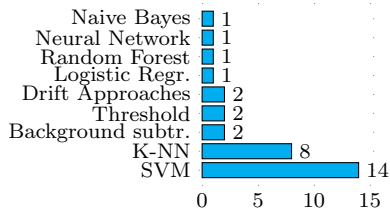
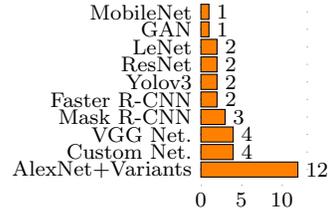
}

Figure \ref{subfig:typesOfFeaturesExtracted} shows that the most popular descriptor is the \ac{LBP} and its variants when considering the feature extraction-based works.
The most popular classifier is the \ac{SVM}, as one can observe in Figure \ref{subfig:numApproachesClassifiers}, and when considering the \ac{DL}-based approaches, the most used network is the AlexNet and the networks derived from it, as shown in Figure \ref{subfig:numApproachesNetwork}.

These findings suggest that research gaps exist for the individual parking spaces classification problem. One research gap is the generalization problem. Future researchers may use multiple datasets to avoid biases, using different datasets for training (e.g., PKLot) and testing (e.g., CNPark-Ext). Standard test protocols considering multiple datasets and standard metrics are also needed to make the comparison between works easier and fairer.

\subsection{Automatic Parking Space Detection}\label{subsec:parkinExtracionDiscussion}

As one can observe in Section \ref{sec:parkingSpacesSegmentation}, only a few automatic parking space detection works were found.
Most discussed approaches did not address the problem, presenting it as an intermediary step for other problems. Except to \citet{padmasiriEtAl2020, kirtibhai2020faster, patel2020car}, no author discussed in Section \ref{sec:parkingSpacesSegmentation} made available quantitative results, making it impossible to evaluate and compare the different works.

For the automatic parking space detection, the \ac{GAN}-based approach proposed in \citet{liChuahBhattacharya2017} may be highlighted.
Although only reporting the individual parking spot classification precision, it generates parking space masks that could be evaluated with appropriate metrics, such as \ac{IoU} or \ac{AP} \citep{everingham2010pascal}.
Other works, such as \citet{vitekMelnicuk2018,nietoEtAl2019}, tried to generate grid-segmentation or fixed-size masks. Nevertheless, a qualitative analysis of the generated parking masks shows that these approaches may be tied to specific parking image properties, such as camera angles and distances.

In \citet{padmasiriEtAl2020} the authors reported results ranging from 59.2 - 63.6 when considering the AP50 metric and 3.16 - 4.75 for the AP75. The authors considered the parking spaces as right angle (90 degrees) bounding boxes, which may not suit angled parking lots, such as the parking lots UFPR04 and UFPR05 of the PKLot dataset. The authors also used images from the same parking lot for both the training and test, which may have led to biased results.

\citet{kirtibhai2020faster, patel2020car} reported automatic parking space detection results using Faster RCNN and YOLOv4 for the car detection step. Then, they detect parking spaces using a car tracking step. Recall/accuracies of 74.5\%/84.79\% and 72.27\%/76.62\%, respectively, were obtained in the CNRPark-EXT dataset. The authors detected right-angle bounding boxes, which may not work in angled parking spaces. Experiments were applied only on three busy days, which may present bias in the test scenario. Also, CNRPark-EXT has only square ground-truths, which do not include the entire parking spaces.

This analysis highlights that automatic parking space segmentation approaches are still needed. It is also imperative that approaches developed for this task present quantitative results. For the quantitative measurement of the results, the authors could use well-known metrics, such as the F1-score, \ac{IoU}, or \ac{AP} \citep{everingham2010pascal}.
The metrics used must not rely on true negatives since the entire parking lot is a true negative, except the parking space polygons.
Metrics that require true negatives, such as accuracy, may lead to biased results.

Standard test protocols are also necessary for this task. Robust test protocols should, for instance, use different datasets for the training and testing phases.
Using different datasets is necessary since approaches developed to automatically extract the parking spaces should perform without any fine-tuning for the specific parking lot.

The human effort to label training samples from the parking lot is more significant than manually marking each parking space. So how can we expect to use labeled images from the test parking lot to train a method developed for this task, as done in \citep{liChuahBhattacharya2017,bohushEtAl2018,padmasiriEtAl2020,kirtibhai2020faster, patel2020car}?
With the same reasoning, methods should not rely on camera parameters, such as the distance from the lens to the parking lot, since these parameters are difficult to acquire in the real world.

\subsection{Car Detection and Counting}\label{subsec:carCountDiscussion}

In this section, a discussion about car detection and counting is presented. Only works that present quantitative results are considered. Overviews of car detection, e.g., detect the bounding boxes of the cars; and counting, e.g., the final number of cars in the images, are showed in Tables \ref{table:comparacaoMetodosDeteccaoCarros} and \ref{table:comparacaoMetodosContagem}, respectively. As done in Section \ref{sec:indivitualParkingSpots}, some results are reported in the $x - y$ format, where $x$ is the worst result achieved in a given scenario, and $y$ is the best result. 

\begin{table}[htpb]
\scriptsize
\setlength{\tabcolsep}{4px}
\caption{Overview of car detection approaches}
\label{table:comparacaoMetodosDeteccaoCarros}
\centering
\begin{threeparttable}[c]
    \begin{tabular}[c]{L{2.2cm}lL{2.2cm}R{2.0cm}l}
        \hline
        Authors & Approach & Metric & Reported Results & Datasets Used\\
        \hline

        \citet{hsiehLinHsu2017} & Custom CNN & \acs{AR} & 57.5 - 62.5\% & PKLot (PUCPR+) \\
        
        \citet{laradjiEtAl2018} & LC-FCN8 & F1-Score & 99\% & PKLot (PUCPR) \\
        
        \citet{vargheseSreelekha2019} & Bag of Features & Precision / Recall & 100\% / 97.2\% & PKLot (single day) \\        

        \citet{liEtAl2019} & Custom CNN & \acs{AP}@.5 and \acs{AP}@.7 & 70.3 - 92.9\% / 44.9 - 61.4\% & PKLot (PUCPR+)\\
        
        \citet{SharmaPandey2021} & Custom CNN & \ac{mAP} & 0.85\tnote{1} & PKLot\\      
        
        \hline
        
    \end{tabular}
    \begin{tablenotes}
        \item[1] The test procedure may have led to biased results or it is not clear. 
    \end{tablenotes}
\end{threeparttable}
\end{table}

\begin{table}[htpb]
\scriptsize
\caption{Overview of car counting approaches.}
\label{table:comparacaoMetodosContagem}
\centering
\begin{threeparttable}[c]
    \begin{tabular}[c]{L{2.2cm}L{2.2cm}lR{2.0cm}l}
        \hline
        Authors & Approach & Metric & Reported Results & Datasets Used\\
        \hline

        \citet{hsiehLinHsu2017} & Custom CNN & \acs{MAE} / \acs{RMSE} & 22.8 - 23.8 / 34.5 - 36.8\tnote{1} & PKLot (PUCPR+)\\

        \citet{laradjiEtAl2018} & LC-FCN8 & \acs{MAE} & 0.2 & PKLot (PUCPR)\\
        
        \citet{amatoEtAl2018, ciampi2018counting} & Mask R-CNN & MAE / RMSE & 1.0 / 2.1 & CNRPark-EXT\\
        
        \citet{stahlEtAl2018} &  R-FCN-based & MAE & 15.1 & PKLot (PUCPR+)\\
        
        \citet{amatoEtAl2019} & Custom CNN & MAE / RMSE & 1.8 - 3.7 / 2.7 - 5.1 & PKLot (PUCPR+) \\

        \citet{liEtAl2019} & Custom CNN & \acs{MAE} / \acs{RMSE} & 3.7 - 9.0 / 5.1 - 9.0 & PKLot (PUCPR+)\\
        
        \citet{dominikGabzdyl2020} & VGG-encoder / Custom CNN & \acs{MAE} / \acs{RMSE} & 7.5 / 8.8 & PKLot (PUCPR+) \\
        
        \citet{YangJia2016} &  Stacked Hourglass (mod.) & MAE / \acs{RMSE} & 2.32 / 3.21 & PKLot (PUCPR+) \\
        
        \hline
        
    \end{tabular}
    \begin{tablenotes}
        \item[1] The test procedure may have led to biased results or it is not clear. 
    \end{tablenotes}
\end{threeparttable}
\end{table}

Some works employed both car detection and counting, appearing in both tables. Since different metrics and datasets were employed when considering different authors in Tables \ref{table:comparacaoMetodosDeteccaoCarros} and \ref{table:comparacaoMetodosContagem}, a direct comparison of the works is unfeasible.

The average \ac{MAE} of the works presented in Table \ref{table:comparacaoMetodosContagem} (\ac{MAE} is the most common metric in Table \ref{table:comparacaoMetodosContagem}) is $7.3$, which is a relatively small error.
However, this result should be considered with care as most of the works employed small subsets of the datasets for the tests.
For instance, the PUCPR+ \citep{hsiehLinHsu2017} subset of the PKLot dataset contains only one day of data.

When considering the datasets, the PUCPR subset of the PKLot dataset and its extended version, PUCPR+ \citep{hsiehLinHsu2017} is popular for the car detection task. It was used in all but two of the reported works in Tables \ref{table:comparacaoMetodosDeteccaoCarros} and \ref{table:comparacaoMetodosContagem}. Many of the presented works, such as  \citet{hsiehLinHsu2017,amatoEtAl2019,dominikGabzdyl2020}, also employed the CARPK \citep{hsiehLinHsu2017}, an interesting dataset containing images captured by a drone. However, this dataset is beyond the scope of this work.

{
    \begin{figure}[htb]
        \centering
        \subfloat[Detection vs. Counting.]{
            \label{subfig:detectVsCount}
            \centering
            
    \tikzsetnextfilename{./tikz/typeTaskCarCount}%
    \begin{tikzpicture}
\small
\toggletrue{showpct}
 \pie [explode=0.05, text=legend, radius=1.0,color={cyan, orange}]
    {62/Detection,
      38/Counting}
\end{tikzpicture}
%

        }
        \subfloat[Classifiers/Features usage.]{
            \label{subfig:methodsUsedCarCount}
            \centering
            
    \tikzsetnextfilename{./tikz/detectorsCarCounting}%
    \begin{tikzpicture}
    \footnotesize,
	\begin{axis}[
		xbar,
		width = 3.2cm,
        height = 3.4cm,
        bar width = 2mm,
		y axis line style = { opacity = 0 },
		tickwidth = 0pt,
		ytick = data,
		symbolic y coords = {Custom CNN,LC-FCN8,Bag of Feat.,Mask R-CNN,R-FCN,VGG,Stacked Hourg.},
		xticklabel={\pgfmathparse{\tick}\pgfmathprintnumber{\pgfmathresult}},
		nodes near coords,
	]
	\addplot[black,fill=cyan] coordinates {
		(7,Custom CNN)
        (2,LC-FCN8)
        (1,Bag of Feat.)
        (1,Mask R-CNN)
        (1,R-FCN)
        (1,VGG)
        (1,Stacked Hourg.)
	};
	\end{axis}
\end{tikzpicture}%

        }
        \caption{Types of tasks and approaches used for the car detection and counting problem. Works that deal with detection and counting tasks may be computed twice.}
        \label{fig:carCountingSummary}
    \end{figure}
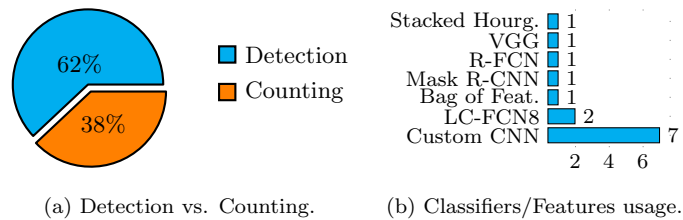
}

Figure \ref{subfig:detectVsCount} shows that 38\% of the approaches may also be used to detect cars. Once the cars are detected in the images, getting the final number of cars becomes a trivial task. Figure \ref{subfig:methodsUsedCarCount} shows the number of works using each strategy. As one can observe, the use of custom \acp{CNN} is the most popular strategy for car detection and counting task.

Similar to the automatic parking space detection task, discussed in Section \ref{subsec:parkinExtracionDiscussion}, there is a lack of common protocols and standard metrics for the car detection and counting task. As discussed in \citet{hsiehLinHsu2017}, another problem in this task is that the datasets must contain a demarcation for all cars present per image in its ground truth. These requirements are not met by any of the datasets discussed in this work. Most authors thus use a subset (often containing all cars manually labeled by the authors of each work) of the original datasets to alleviate this problem. This approach is suboptimal since the subsets are often small, e.g., one day of images, and may not represent the possible conditions encountered in the real world. It indicates that the datasets' ground-truth discussed in this work could be improved by adding all cars' positions in the ground-truth.

\section{Conclusion and Future Works}\label{sec:conclusion}

This paper presented a systematic review of the vision-based approaches that address parking lot management problems. To the best of our knowledge this is the first such vision-centered review.
First, considering the reproducibility of results, the public datasets related to vision-based parking lot management were analyzed. Criteria were established based on which the available datasets were filtered to obtain those relevant to this problem. Three image-based datasets passed the established criteria: PKLot, CNRPark-EXT, and PLds datasets. When considering the reviewed works that use these datasets, we concluded that the PKLot is the most popular dataset, used in \pklotOverallUsagePct\ of the surveyed works. This was followed by the CNRPark-EXT and PLDs datasets, employed by \cnrparkOverallUsagePct\ and \pldsOverallUsagePct\ of the works, respectively.

Although we consider both image and video-based datasets relevant in this work, we found that only the image-based datasets met our restrictions. Thus, robust video-based parking lot datasets can be a future contribution to the scientific community. Video-based datasets could be helpful in various ways, for instance, tracking cars or to identify suspicious behavior. In addition, only a few samples from the analyzed datasets were from night-time and or had snow weather. Thus, datasets containing more of these and other scenarios may be a contribution to future research. Further, datasets containing labels, for example, bounding boxes, for all cars available in each image could also be interesting for future developments, particularly for car counting.

When considering the works that used the surveyed datasets, we discovered that the authors focused on three main tasks: the individual parking spaces classification (between occupied and empty), the automatic parking spaces detection, that is, automatically detect the parking space locations in the images, and the car detection and counting, for example, counting the number of vehicles in the images.

Overall, \overallParkingClassificationOnly\ of the reviewed works proposed approaches exclusively to address the individual parking spaces classification problem. However, this task has been well-addressed considering that both the train and test images originate from the same parking lot and camera angle, reaching accuracy rates of \globalAverageAccSinglePK\, on average. However, we concluded that only \numWorksIndividualClassConsiderChange\ of the works that deal with the individual parking spaces classification consider parking lot or camera angle changes.

On average, the accuracy reached in the surveyed works in scenarios containing camera angle and parking lot changes was \globalAverageAccCamChange\ and \globalAverageAccPkChange, respectively. Thus, with this in mind, future works should focus on camera and parking lot changes, that is, scenarios where the training images originate from a parking lot or camera angle different from the test images.

Further, only \globalParkingSpotDetectionWorks\ of the surveyed works dealt with the automatic parking spot detection problem, and only three authors made the quantitative results available. This absence of quantitative results can act as an encouragement for significant research efforts for this task. In addition, by reducing the human labor in labeling the parking spaces positions in new parking lots (or when the camera angle changes), automatic parking space detection approaches could lead to more robust and easier to deploy vision-based systems crafted to monitor parking spaces areas.

Of all the surveyed works, \globalCarDetecCountWorks\ considered the car detection and counting task. Most of these works used the PUCPR subset of the PKLot dataset for the tests. Many of these approaches prefer using image datasets from non-fixed cameras, for instance, ones using drones, such as the CARPK \citep{liEtAl2019}. However, although interesting, these datasets are beyond the scope of this work.

There is a lack of standard protocols for testing approaches for all tasks discussed in this work, rendering the comparison of works and reproduction of the experiments a challenge. Future test protocols should consider which evaluation metrics should be used, the manner in which the data should be split, which datasets exhibit which challenges, and so forth. The proposal and usage of well-defined test protocols could provide better insights and lead to more robust developments.

When comparing handcrafted feature descriptors and deep learning-based approaches, 35\% of the surveyed works employed the former, and 61\% the latter (4\% of the methods used both). 
For the individual parking spaces classification problem, the method suited to obtain results in not clear. However, on average, deep learning methods tend to generate slightly better accuracies. Moreover, it is crucial to notice that most works lack information about, for instance, the computational cost of the proposed methods. High computational costs could result in certain approaches being prohibitive for certain applications, such as in embedded systems. For the car detection and counting task, all but one of the surveyed approaches employed deep learning-based approaches, clearly indicating that the authors prefer this technique.

Moreover, we also want to point out that many authors included small private datasets in the tests, despite the availability of public datasets. Although we recommend the development and usage of new datasets, the usage of small datasets may lead to biased results. In addition, by making these datasets private, authors are undermining the scientific community’s ability to reproduce the results. Further, authors should also consider the specific properties of parking lot image datasets, such as the occurrence of a specific car in several images, to avoid trivial and unrealistic test protocols.

We hope that new research works and datasets can be developed based on the points discussed in this work. As made evident in this work, new developments should consider making the train and test data publicly available. These proposals would improve experiment reproducibility and contribute to the scientific community with new datasets.

\biboptions{authoryear}
\bibliography{references}

\end{document}